\documentclass[sigconf]{acmart}

\AtBeginDocument{%
  }

\copyrightyear{2026}
\acmYear{2026}
\setcopyright{cc}
\setcctype{by}
\acmConference[CIKM '26]
  {Proceedings of the 35th ACM International Conference on Information and Knowledge Management}
  {November 7--11, 2026}
  {Rome, Italy.}
\acmBooktitle{Proceedings of the 35th ACM International Conference on Information and Knowledge Management (CIKM '26), November 7--11, 2026, Rome, Italy}
\acmISBN{979-8-4007-2539-5/2026/11}
\acmDOI{10.1145/3799682.3840928}

\usepackage{graphicx}
\graphicspath{{figures/}}
\usepackage{tabularx}
\usepackage{array}
\usepackage{placeins} % \FloatBarrier
\usepackage{ragged2e}
\usepackage{microtype}
\usepackage{algorithm}
\usepackage{algpseudocode}
\usepackage{enumitem}
\usepackage{makecell}
\usepackage{multirow}
\usepackage{balance}
\newcolumntype{Y}{>{\centering\arraybackslash}X}
\begin{document}

\title{GroupSegment-SHAP: Shapley Value Explanations with Group-Segment Players for Multivariate Time Series}

\author{Jinwoong Kim}
\orcid{0009-0001-0707-5908}
\affiliation{%
  \department{Department of Industrial Data Engineering}
  \institution{Hanyang University}
  \city{Seoul}
  \country{Republic of Korea}
}
\email{dnddl9456@hanyang.ac.kr}

\author{Sangjin Park}
\authornote{Corresponding author.}
\orcid{0009-0008-6922-624X}
\affiliation{%
  \department{Department of Industrial Data Engineering}
  \institution{Hanyang University}
  \city{Seoul}
  \country{Republic of Korea}
}
\email{psj3493@hanyang.ac.kr}

\renewcommand{\shortauthors}{Jinwoong Kim and Sangjin Park}

\begin{abstract}
Multivariate time-series models achieve strong predictive performance in healthcare, industry, energy, and finance, but how they combine cross-variable interactions with temporal dynamics remains unclear. SHapley Additive exPlanations (SHAP) has been widely used for interpretation. However, existing time-series variants typically treat the feature and time axes independently, fragmenting structural signals formed jointly by multiple variables over specific intervals. We propose GroupSegment-SHAP (GS-SHAP), which constructs explanatory units as group-segment players based on cross-variable dependence and distribution shifts over time, and then quantifies each unit's contribution via Shapley attribution. We evaluated GS-SHAP across four real-world domains: human activity recognition, power-system forecasting, medical signal analysis, and financial time series, and compared it with KernelSHAP, TimeSHAP, SequenceSHAP, WindowSHAP, and TSHAP. GS-SHAP consistently achieves the highest deletion-based faithfulness ($\Delta$AUC) across all four benchmark datasets.
In synthetic evaluations with controlled ground-truth structures, GS-SHAP achieves 24.4\% higher IoU-based recovery of multivariate-temporal patterns than the strongest baseline. In addition, GS-SHAP runs on average 3.38$\times$ faster than TimeSHAP across approximation budgets in power-system forecasting, showing that it can simultaneously achieve high explanatory faithfulness and computational efficiency. In a financial case study, GS-SHAP identifies interpretable multivariate-temporal interactions among key market variables across market regimes, highlighting its potential utility for risk-aware investment analysis. Code is available at \url{https://github.com/jirungx/GroupSegment-SHAP}.
\end{abstract}




\maketitle

\section{Introduction}

Multivariate time-series prediction is crucial in many real-world domains, such as healthcare monitoring, industrial process control, energy management, and financial analysis \cite{RN3,RN2,RN1}. Because multiple variables co-evolve and interact over time, capturing these dependencies is essential for predictive accuracy and reliable decision-making. Recent deep learning models have substantially improved performance by learning such complex patterns \cite{RN5,RN4}. In particular, Recurrent Neural Network (RNN)-based architectures such as LSTM \cite{hochreiter1997long} and GRU \cite{cho2014learning} remain widely used for modeling sequential data. However, their black-box nature limits transparency, making it difficult to interpret why specific predictions are made \cite{RN7,RN6}. This is particularly problematic in real-time, high-stakes applications, including medical anomaly detection, equipment failure diagnosis, energy demand forecasting, and financial risk monitoring, where explanations should reveal how predictions arise from intervariable interaction patterns and coupled temporal dynamics \cite{RN7,RN9,RN8}.

To address this issue, explainable AI methods such as LIME and SHAP have been developed \cite{RN69,RN13}. LIME explains an individual prediction by fitting a local surrogate around the input. SHAP, based on Shapley values from cooperative game theory, attributes each feature's contribution under a consistent axiom set \cite{RN69,RN13}. SHAP is also model agnostic and provides directly interpretable attribution scores, making it a widely adopted standard across domains such as mobile sensing \cite{RN53}, energy \cite{RN52}, healthcare \cite{RN51}, and finance \cite{RN72,RN73}. These properties have motivated recent extensions of SHAP to time series to interpret the temporal structure itself \cite{RN12,RN10}.

\begin{figure}[h]
  \centering
  \includegraphics[width=0.95\linewidth]{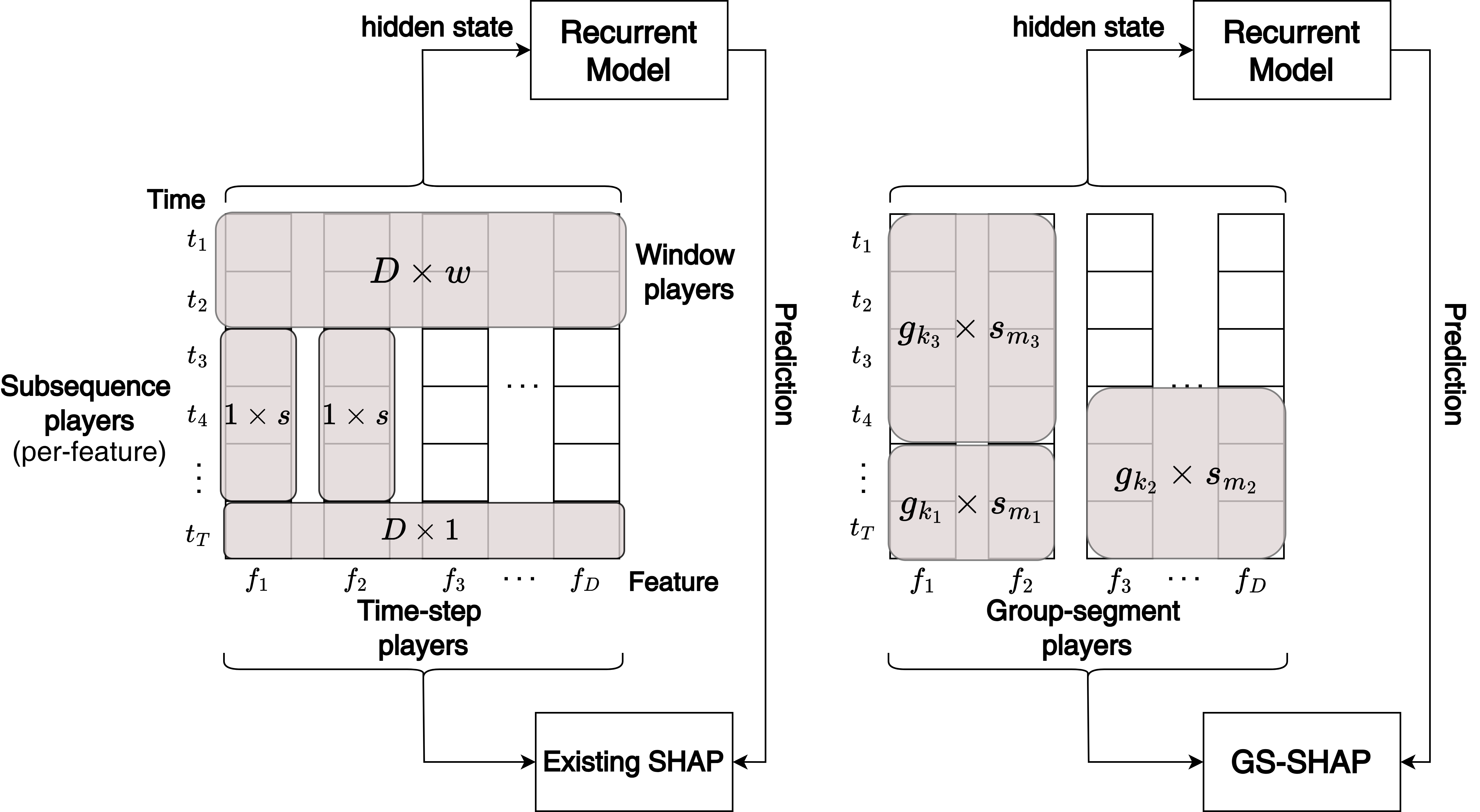}
  \caption{Comparison of player designs in existing sequential SHAP variants and GS-SHAP.}
  \label{fig:intro_overview}
\end{figure}

Several SHAP-based explainers have been proposed for time-series models, including variants that define players as individual cells, time steps, fixed windows, or temporal subsequences and estimate attributions via perturbation-based model queries \cite{RN13,RN12,RN10}. However, these player definitions largely preserve a single-axis view of the input, either aggregating over time while treating variables independently or assigning importance to fine-grained cells. As a result, the Shapley game can be defined over players that do not match the multivariate-temporal evidence used by the predictor. Figure~\ref{fig:intro_overview} contrasts these player designs with the joint group-segment players used in GS-SHAP.

Existing explainers often decouple temporal and feature axes, hindering the representation of structurally meaningful multivariate-temporal interactions as coherent units in time series \cite{RN17,RN16}. In many settings, the most meaningful evidence arises when groups of variables change concurrently over specific intervals, and such joint patterns frequently constitute the model's true predictive rationale \cite{RN15,RN14}. Prior methods tend to fragment or distort this coupled structure, fundamentally limiting the ability to structurally recover the signals used by the model \cite{RN17,RN16}. This limitation is particularly consequential in practical domains such as healthcare, industry, energy, and finance, where explanation-driven decision-making is often required, motivating explanatory units that naturally encode coupled spatiotemporal structure \cite{RN7,RN8,RN19,RN18}.

We propose GroupSegment-SHAP (GS-SHAP), a SHAP-based framework that redefines the attribution player space for multivariate time series. Rather than treating features, timestamps, or fixed temporal windows as separate players, GS-SHAP defines each player as the intersection of a dependency-based feature group and a distribution-shift-based temporal segment. This makes coupled multivariate-temporal patterns the atomic units of the Shapley game, allowing attributions to be assigned to coherent structures rather than being dispersed across disconnected cells or single-axis segments. Across four heterogeneous domains, including UCI Human Activity Recognition, power-system forecasting, electrocardiogram signal analysis, and financial time series, GS-SHAP achieves stronger faithfulness and more stable attribution patterns than existing explainers.

Overall, we summarize our contributions in three key points as follows:
\begin{itemize}
  \item Group-segment players are introduced as a new Shapley player space that addresses player-space mismatch by representing structurally coherent variable-time regions in multivariate time series.
  \item GS-SHAP provides a general player-construction and attribution framework that reduces attribution fragmentation caused by feature-only, time-only, or fixed-window player definitions.
  \item GS-SHAP provides a faithful, stable, and computationally efficient explanation framework for heterogeneous multivariate time series, enabling reliable attribution across diverse real-world domains and controlled synthetic benchmarks.
\end{itemize}

The paper is organized as follows. Section~2 reviews related work; Section~3 presents the proposed method; Section~4 reports experimental results; Section~5 presents an S\&P500 market-regime case study; and Section~6 concludes.

\section{Related Work}

\subsection{Explainability in Multivariate Prediction}
Multivariate time-series prediction models achieve strong performance across diverse real-world settings \cite{RN2,RN1,RN21,RN20} by learning patterns from jointly evolving variables. RNN-based forecasters are widely used in practice, yet prior work cautions that off-the-shelf explainers can under-attribute past events and long-range dynamics while over-emphasizing the current input \cite{RN12}. For example, many time-series adaptations define players on a single axis (e.g., time steps, fixed windows) or flatten the input into independent cells, which can distort multivariate-temporal structure \cite{RN17,RN16}. However, it remains unclear how variable groups form spatiotemporal interactions within specific temporal regimes during prediction \cite{RN8,RN17,RN16}. Existing explainers often decompose only the feature or temporal axis, making it difficult to recover the interaction structure used by the model \cite{RN17,RN16}.

These limitations are particularly salient in high-stakes domains. In healthcare, the joint activation of physiological indicators can represent clinically meaningful risk signals \cite{RN28,RN27}. In industrial equipment, coupled patterns across sensors may constitute precursor signals of failures \cite{RN26,RN25}. In financial markets, coupled variations of price, trading volume, and volatility over specific intervals can indicate market regime transitions \cite{RN24,RN23}. In such environments, importance scores at the level of a single feature or time point are insufficient to capture the structural basis of model decisions \cite{RN17,RN16}. Therefore, new approaches are required that incorporate spatiotemporal interactions induced by cross-variable dependence and temporal dynamics into the explanatory unit (player) for multivariate time series \cite{RN8,RN10,RN15}.

\vspace{-1.0em}

\begin{figure*}[h]
  \centering
  \includegraphics[width=0.82\linewidth]{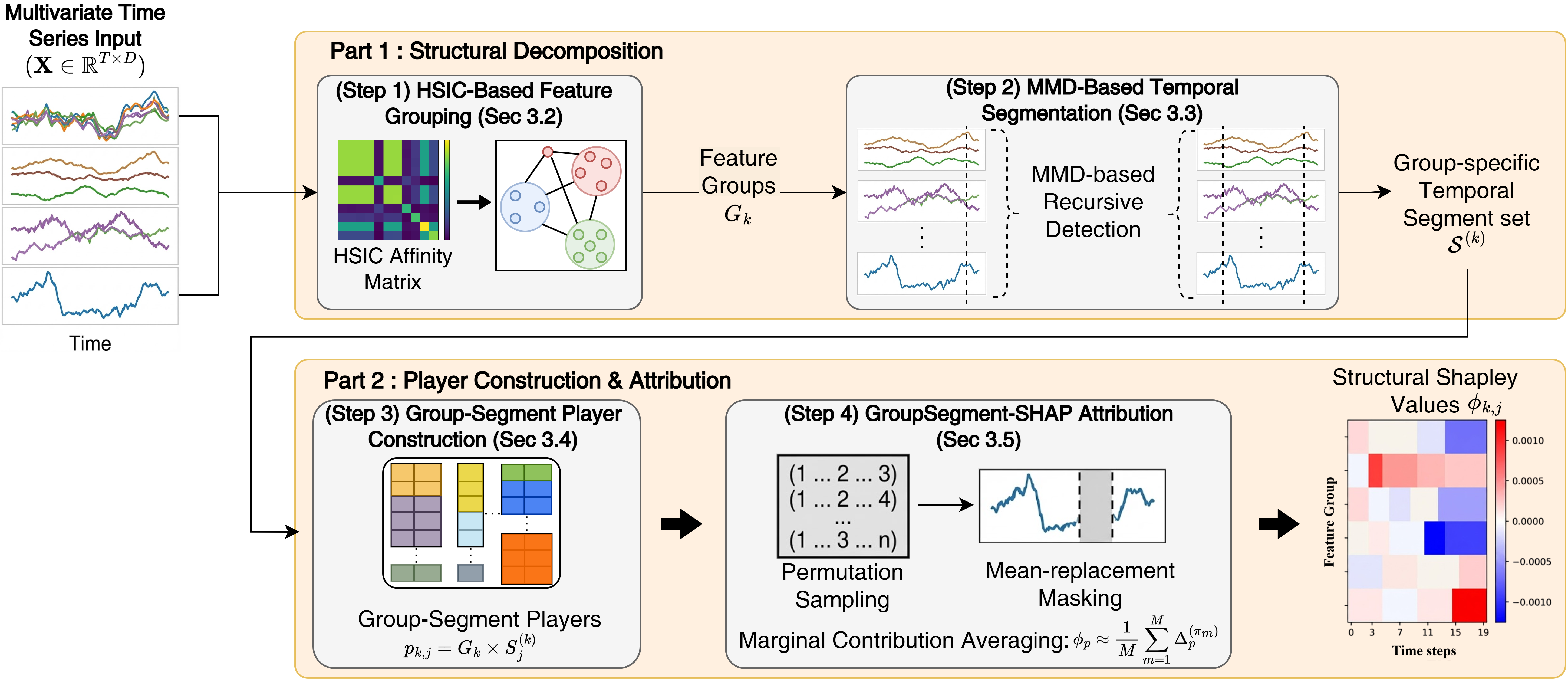} 
  \caption{Overview of the GS-SHAP framework.}
  \label{fig:overview}
\end{figure*}

\subsection{SHAP Explanations for Sequential Data}
Shapley value-based methods are widely used to interpret time-series models because they provide axiomatic attributions that quantify each input's contribution to model-output changes \cite{RN12,RN15,RN11}. However, KernelSHAP \cite{RN13} treats inputs as a static set of feature-wise players and its perturbation-based coalitions do not reflect sequential structures such as temporal continuity, intertemporal dependence, and cross-variable co-activation \cite{RN12,RN11,RN29}. This player-space mismatch becomes more problematic in high-dimensional multivariate time series, where predictive evidence often arises from spatiotemporal interaction patterns rather than isolated features or cells \cite{RN17,RN16,RN30}.

To account for temporal structure, TimeSHAP \cite{RN12} estimates time-step-level contributions through event-level perturbations, yielding intuitive temporal importance scores. However, it explains time points rather than structured units that couple multiple variables \cite{RN15}. Cell-level explanations provide finer granularity, but they often disperse multivariate interactions across many variable-by-time cells, weakening the structural meaning of joint spatiotemporal patterns \cite{RN10,RN31}.

For longer temporal patterns, SequenceSHAP \cite{RN10} introduces temporal segmentation and estimates subsequence-level importance. Although this helps identify extended temporal evidence, it applies common segments across all variables and treats the feature axis independently \cite{RN10}, which can be restrictive when dependencies are localized to specific feature subsets. Similarly, WindowSHAP \cite{RN11} and TSHAP \cite{RN35} define contiguous temporal windows as players, but their time-centric aggregation does not explicitly model cross-variable coupling or data-driven heterogeneity in temporal regimes. ShaTS \cite{de2025shats} incorporates \emph{a priori} grouping to produce more coherent attributions, but relies on predefined group structures rather than data-driven spatiotemporal explanation units.

Overall, existing SHAP-based explainers often define players along a single axis, limiting their ability to capture the spatiotemporal interaction structure used by multivariate sequential models \cite{RN12,RN10,RN11,RN31,RN35}. This limitation can undermine explanation-driven decision-making \cite{RN7,RN8,RN19,RN18} and motivates Shapley player spaces that jointly reflect cross-variable dependence and temporal dynamics.

\vspace{-0.8em}

\section{Methodology}

Figure~\ref{fig:overview} summarizes GS-SHAP for a multivariate time series $X\in\mathbb{R}^{T\times D}$ in four steps:
\begin{enumerate}[
  label=\textbf{(Step~\arabic*)},
  leftmargin=*,
  labelsep=0.6em,
  align=left,
  itemsep=2pt,
  topsep=2pt
]
  \item \textbf{HSIC-Based Feature Grouping:} Partition the variable space into feature groups based on nonlinear dependencies among variables.
  \item \textbf{MMD-Based Temporal Segmentation:} Detect distributional changes in the time series and derive temporal segments.
  \item \textbf{Group-Segment Player Construction:} Define group \-segment players in the form of feature-group time intervals by combining the resulting feature groups and temporal segments.
  \item \textbf{GroupSegment-SHAP Attribution:} Approximate each player's marginal contribution using Shapley values and compute attributions.
\end{enumerate}

This structural decomposition mitigates attribution fragmentation from feature-level, time-step-level, or fixed-window-level SHAP explanations, enabling more consistent interpretation of the multivariate-temporal structures used by the model.

\subsection{Problem Definition}
We study a multivariate time series with length $T$ and variable dimension $D$. The observation at time step $t$ is defined as follows:
\begin{equation}
x_t \in \mathbb{R}^{D}.
\label{eq:1}
\end{equation}
The full sequence is obtained by aggregating observations over time.
\begin{equation}
X = \{x_t\}_{t=1}^{T} \in \mathbb{R}^{T \times D}.
\label{eq:2}
\end{equation}
Here, we assume a black-box predictor $f:\mathbb{R}^{T \times D}\rightarrow \mathbb{R}$ that outputs a scalar prediction $\hat{y}=f(X)$. The goal is to quantify which multivariate-temporal structures in $X$ contribute to $\hat{y}$. Existing SHAP-based time-series explainers typically define players as individual features or time steps, which distribute joint multivariate-temporal patterns across multiple players and produce fragmented attribution signals, leading to the fragmentation issue \cite{RN10,RN31}. To address this, GS-SHAP derives feature groups based on nonlinear intervariable dependence and temporal segments driven by distribution shifts, then combines them into group-segment players. This preserves each multivariate-temporal structure as a single explanatory unit and reveals coupled temporal dynamics that are difficult to capture with feature- or timestamp-level explanations.

\subsection{HSIC-Based Feature Grouping}
Multivariate time series often exhibit nonlinear cross-variable dependencies, and predictive signals frequently arise from joint patterns spanning multiple variables \cite{RN40,RN39}. Variable-wise explanation units can fragment such coupled structures, motivating the grouping of strongly interdependent variables into shared interpretation units \cite{RN38}. 

We construct feature groups using the Hilbert-Schmidt independence criterion (HSIC) \cite{RN37,RN36}. HSIC measures statistical dependence, not causality; thus, the resulting feature groups should not be interpreted as causal groups. Specifically, we build an HSIC affinity matrix, select the number of groups via the eigengap criterion, and apply spectral clustering to obtain $G=\{G_1,\ldots,G_K\}$. These groups serve as the multivariate structural units for group-segment players.

\subsubsection{Measuring Nonlinear Dependency using HSIC}
HSIC is a kernel-based dependence measure that captures high-order nonlinear dependency beyond linear correlation \cite{RN37}. Let $n$ denote the number of observations used to estimate HSIC. The centering matrix $H\in\mathbb{R}^{n\times n}$ is
\begin{equation}
H = I_n - \frac{1}{n}\mathbf{1}\mathbf{1}^{\top},
\label{eq:3}
\end{equation}
where $I_n$ is the $n\times n$ identity matrix and $\mathbf{1}\in\mathbb{R}^{n}$ is the all-ones vector. For two variables (dimensions) $X_d$ and $X_{d'}$ with a Gaussian (RBF) kernel, HSIC is computed as
\begin{equation}
\mathrm{HSIC}(X_d, X_{d'}) = \frac{1}{(n-1)^2}\,\mathrm{tr}(H\mathbf{K}_dH\mathbf{K}_{d'}),
\label{eq:4}
\end{equation}
where $\mathbf{K}_d,\mathbf{K}_{d'}\in\mathbb{R}^{n\times n}$ are kernel matrices for $X_d$ and $X_{d'}$, and $\mathrm{tr}(\cdot)$ denotes the trace operator. We set the RBF bandwidth using the median heuristic on the sampled values for each variable pair. Here $N$ is the number of training sequences in $\mathcal{D}_{\mathrm{train}}$, and $t$ indexes time steps within each sequence. To estimate global dependence, we construct per-variable samples from the training set $\mathcal{D}_{\mathrm{train}}=\{X^{(i)}\}_{i=1}^{N}$ by collecting $\{X^{(i)}_{t,d}\}_{i,t}$ for each variable $d$. 

For computational efficiency, HSIC is estimated using a fixed-size random subsample of these observations, and the resulting affinity is reused across all explanation runs. Larger HSIC indicates stronger nonlinear dependence and is used to define feature groups \cite{RN37,RN49}. Localized distribution shifts are addressed in the subsequent temporal segmentation stage.

\subsubsection{Determining the Number of Groups via Eigengap and Creating Feature Groups}
Let $D$ denote the number of variables. Given the HSIC affinity matrix $A\in\mathbb{R}^{D\times D}$, we define the normalized graph Laplacian
\begin{equation}
D_A=\mathrm{diag}(A\mathbf{1}), \qquad
L_A = I_D - D_A^{-1/2}AD_A^{-1/2},
\label{eq:5}
\end{equation}
where $\mathbf{1}\in\mathbb{R}^{D}$ is the all-ones vector, $D_A$ is the degree matrix, $\mathrm{diag}(\cdot)$ maps a vector to a diagonal matrix, and $I_D$ is the $D\times D$ identity matrix. We choose $K$ by the eigengap in the eigenvalues of $L_A$, embed variables using the top $K$ eigenvectors of $L_A$, and apply $k$-means to obtain $G_1,\ldots,G_K$. Computing $A$ costs $O(D^2 n^2)$, but it is a one-time preprocessing step on training data and is reused across all explanation runs.

\subsection{MMD-Based Temporal Segmentation}
Multivariate time series often preserve statistical characteristics over an interval and then exhibit abrupt distribution shifts at specific time points \cite{RN41,RN42}. Such regime shifts are critical for predictions in settings involving behavioral changes, transitions in physiological patterns, or market events. Thus, partitioning the time axis with fixed intervals or predefined windows may fail to reflect the underlying temporal structure \cite{RN43}. 

To obtain data-driven temporal segments, we detect distribution shifts using the maximum mean discrepancy (MMD) \cite{RN44,RN45}. Given an interval $[s,e)$, for each candidate split $t\in(s,e)$, we define the left and right segments $X_L=X[s:t)$ and $X_R=X[t:e)$, respectively, and compute their discrepancy using the unbiased MMD estimator.
\begin{align}
\raisetag{6pt}
\mathrm{MMD}^2(X_L,X_R)
&=
\frac{1}{n(n-1)}\sum_{i\neq i'} k(x_i,x_{i'})
+\frac{1}{m(m-1)}\sum_{j\neq j'} k(y_j,y_{j'}) \nonumber \\
&-
\frac{2}{nm}\sum_{i=1}^{n}\sum_{j=1}^{m} k(x_i,y_j),
\label{eq:mmd}
\end{align}
where $\{x_i\}_{i=1}^{n}$ and $\{y_j\}_{j=1}^{m}$ are samples from $X_L$ and $X_R$, and $n,m$ denote the segment lengths. MMD captures distributional differences beyond mean shifts, including changes in variance, correlation structure, and nonlinear dependencies, making it suitable for regime shift detection. If the MMD value at a split exceeds a threshold $\tau$, we accept $t$ as a change point and recursively apply the same search to the two subintervals $[s,t)$ and $[t,e)$. Repeating this procedure until no further change points are detected yields a data-driven segmentation of the interval. We set $\tau$ following standard kernel two-sample testing practice by approximating the null distribution via permutations and selecting the upper quantile at significance level $\alpha$ \cite{RN22,RN45,RN55}.

To account for heterogeneous change patterns across feature groups, GS-SHAP applies the same shift-detection procedure independently to each feature group $G_k$, producing a group-specific set of temporal segments $\mathcal{S}^{(k)}$. For group $k$, we denote the segmentation as $\mathcal{S}^{(k)}$, a collection of nonoverlapping segments along the time axis.
\begin{equation}
\mathcal{S}^{(k)}=\{\,S_1^{(k)}, S_2^{(k)}, \ldots, S_{J_k}^{(k)}\,\},\qquad
S_j^{(k)}=\bigl[t_{j-1}^{(k)},\,t_j^{(k)}\bigr),
\label{eq:segset}
\end{equation}
where $J_k$ is the number of segments for group $k$, and each $S_j^{(k)}$ denotes a disjoint interval. To ensure comparability and segmentation stability, we use a common set of segmentation hyperparameters across all feature groups. Specifically, we select the Gaussian kernel bandwidth using the median heuristic, enforce a minimum segment length $L_{\min}$ to avoid unstable over-segmentation, set the change-point threshold $\tau$ according to the significance level of the permutation-based two-sample test, and cap the maximum number of segments by $J_{\max}$ to prevent noise-driven fragmentation. Unlike HSIC grouping, which is estimated once from the training set and reused across explanations, MMD-based segmentation is performed on each explained input sequence within each feature group; its cost is included in the end-to-end runtime reported in Section~\ref{sec:computational_efficiency}.

\subsection{Group-Segment Player Construction}
Feature grouping yields variable groups $G=\{G_1,\ldots,G_K\}$. For each group $G_k$, MMD-based detection produces a group-specific set of temporal segments $\mathcal{S}^{(k)}$. GS-SHAP combines these variable groups and temporal segments to define group-segment players.

Each group-segment player is treated as an independent explanatory unit for Shapley coalition construction. Here, $G_k$ preserves multivariate interactions among jointly varying variables, and $S_j^{(k)}\in \mathcal{S}^{(k)}$ denotes an individual temporal segment for group $k$. This coupling preserves a multivariate-temporal pattern as a single unit of interpretation. For each pair of $G_k$ and $S_j^{(k)}$, the corresponding multivariate subsequence is defined as follows:
\begin{equation}
X_{(G_k,S_j^{(k)})}
=
\left\{\, X_{t,d}\mid d\in G_k,\; t\in S_j^{(k)} \,\right\},
\label{eq:8}
\end{equation}
where $X_{t,d}$ is the observation of variable $d$ at time $t$. The complete set of group-segment players is defined as follows:
\begin{equation}
P
=
\left\{\, p_{k,j}\mid k=1,\ldots,K;\; j=1,\ldots,J_k \,\right\},
\qquad
|P|
=
\sum_{k=1}^{K} J_k .
\label{eq:9}
\end{equation}

A key property is that the group-segment players form a non-overlapping partition of the input grid. Each variable $d\in\{1,\ldots,D\}$ belongs to a unique group $G_k$, and $\mathcal{S}^{(k)}$ partitions the time axis $\{1,\ldots,T\}$ into disjoint segments. Therefore, each cell $(t,d)$ is assigned to a unique pair $(k,j)$ such that $d\in G_k$ and $t\in S_j^{(k)}$. This uniqueness prevents overlaps or conflicts in subsequent masking and cell-level importance aggregation, while defining a valid finite player set on which the standard Shapley axioms remain applicable at the group-segment level.

\subsection{GroupSegment-SHAP Attribution}
Once group-segment players are defined, we quantify each structural unit's contribution to the model output using Shapley values \cite{RN13,RN46}. Let $P$ denote the set of players and $p\in P$ a target player. The Shapley value of $p$ is
\begin{equation}
\phi_p
=
\sum_{C\subseteq P\setminus\{p\}}
\frac{|C|!\,(|P|-|C|-1)!}{|P|!}
\left[
f\!\left(\tilde{X}^{(C\cup\{p\})}\right)
-
f\!\left(\tilde{X}^{(C)}\right)
\right],
\label{eq:10}
\end{equation}
where $C$ is a coalition of players, $\{p\}$ is the singleton set containing $p$, and $P\setminus\{p\}$ denotes the remaining players excluding $p$. The function $f(\cdot)$ is the predictive model, and $\tilde{X}^{(C)}$ is a masked input that activates only the group-segment regions included in $C$ (all others are masked). Since evaluating all subsets is infeasible, we approximate Shapley values via permutation sampling. Let $\pi$ be a random permutation of $P$ and define the set of players appearing before $p$ in $\pi$ as
\begin{equation}
\mathrm{Pre}_{\pi}(p)
=
\left\{\, q\in P \mid q \text{ appears before } p \text{ in } \pi \,\right\}.
\label{eq:11}
\end{equation}
Under $\pi$, the marginal contribution of $p$ is
\begin{equation}
\Delta_{p}^{(\pi)}
=
f\!\left(\tilde{X}^{(\mathrm{Pre}_{\pi}(p)\cup\{p\})}\right)
-
f\!\left(\tilde{X}^{(\mathrm{Pre}_{\pi}(p))}\right),
\label{eq:12}
\end{equation}
and we estimate $\phi_p$ by averaging over $M$ sampled permutations $\{\pi_m\}_{m=1}^{M}$:
\begin{equation}
\phi_p
\approx
\mathbb{E}_{\pi}\!\left[\Delta_{p}^{(\pi)}\right]
\approx
\frac{1}{M}\sum_{m=1}^{M}\Delta_{p}^{(\pi_m)}.
\label{eq:13}
\end{equation}

Shapley computation requires masking players excluded from a coalition. We adopt mean-replacement masking with a per-feature baseline $\mu_d$. For consistency with our experiments, $\mu_d$ is computed as the feature-wise mean over the same background set used in comparative evaluations. In addition to mean replacement, we also evaluate alternative masking baselines (zero and noise) under the same perturbation protocol.

We represent a coalition $C\subseteq P$ by a binary vector $z\in\{0,1\}^{|P|}$, where $z_p=1$ indicates that player $p$ is active and $z_p=0$ indicates that it is masked. Let $(t,d)$ index time and feature dimensions, and let $p(t,d)\in P$ denote the unique player that contains cell $(t,d)$. The masked input $\tilde{X}^{(z)}$ is defined element-wise as
\begin{equation}
\tilde{X}^{(z)}_{t,d}
=
\begin{cases}
X_{t,d}, & z_{p(t,d)} = 1,\\
\mu_d,   & z_{p(t,d)} = 0.
\end{cases}
\label{eq:14}
\end{equation}
Thus, included regions retain original values while excluded regions are replaced by $\mu_d$. The overall GS-SHAP procedure is summarized in Algorithm~\ref{alg:gs_shap}, and its stage-wise theoretical complexity is provided in Appendix~\ref{app:complexity_analysis}.

\begin{algorithm}[h]
\caption{GroupSegment-SHAP Procedure}
\label{alg:gs_shap}
% \small
\begin{algorithmic}[1]
\Require Time series $X\in\mathbb{R}^{T\times D}$, trained model $f$, background set $\mathcal{B}$, permutation budget $M$
\Ensure Group-segment Shapley values $\{\phi_p\}_{p\in P}$

\State Compute the HSIC affinity matrix $A$ for variable pairs (Eq.~\ref{eq:4}).
\State Determine $K$ and feature groups $G=\{G_1,\ldots,G_K\}$ by spectral clustering on $A$ (Eq.~\ref{eq:5}).
\State Compute the feature-wise masking baseline $\mu$ from $\mathcal{B}$ (Eq.~\ref{eq:14}).
\For{each feature group $G_k$}
    \State Detect change points using MMD-based recursive segmentation (Eq.~\ref{eq:mmd}).
    \State Obtain group-specific segments $\mathcal{S}^{(k)}=\{S^{(k)}_1,\ldots,S^{(k)}_{J_k}\}$ (Eq.~\ref{eq:segset}).
\EndFor
\State Construct group-segment players $P=\{p_{k,j}:G_k\times S^{(k)}_j\}$ (Eqs.~\ref{eq:8}--\ref{eq:9}).
\For{$m=1$ to $M$}
    \State Sample a random permutation $\pi_m$ of players in $P$.
    \For{each player $p\in P$}
        \State Identify preceding players $\mathrm{Pre}_{\pi_m}(p)$ (Eq.~\ref{eq:11}).
        \State Compute the marginal contribution $\Delta_p^{(\pi_m)}$ with mean-replacement masking (Eqs.~\ref{eq:12} and ~\ref{eq:14}).
    \EndFor
\EndFor
\State Estimate $\phi_p$ by averaging $\Delta_p^{(\pi_m)}$ over $M$ permutations (Eq.~\ref{eq:13}).

\end{algorithmic}
\end{algorithm}

\section{Experiments}
\subsection{Experimental Setup}

\textbf{Datasets and Prediction Tasks.}
GS-SHAP is evaluated on four multivariate time-series datasets spanning mobile sensing, energy, healthcare, and finance: HAR, ETTm1, PTB-XL, and S\&P500. These datasets support the main evaluations of explanation quality, feature grouping, robustness, and runtime. All tasks use a fixed-length input window $T$, with data sources and preprocessing following the original papers and public repositories.

\begin{itemize}
  \item \textbf{HAR} is the UCI-HAR six-class activity recognition benchmark with nine inertial variables collected from a waist-mounted smartphone \cite{RN53}.
  \item \textbf{ETTm1} is a 15-min transformer-operation benchmark for 1-h-ahead load prediction using seven variables \cite{RN52,RN60,RN59}.
  \item \textbf{PTB-XL} provides 10-s 12-lead ECG waveforms, from which we construct a binary normal-versus-abnormal classification task \cite{RN51}.
  \item \textbf{S\&P500} contains eleven financial variables, including OHLCV, SMA10/20, VIX, DXY, WTI, and Gold, for next-trading-day return prediction \cite{RN58}.
\end{itemize}

Table~\ref{tab:window} summarizes the benchmark settings. The minimum segment length $L_{\min}$ is set proportional to $T$, with a lower bound for short sequences, to avoid unstable over-segmentation \cite{RN51,RN20}.

\begin{table}[h]
\centering
\caption{Summary of main benchmark datasets and experimental settings.}
\label{tab:window}
\setlength{\tabcolsep}{3pt}
\renewcommand{\arraystretch}{1.05}
\footnotesize
\begin{tabular*}{\linewidth}{@{\extracolsep{\fill}}lcccc}
\hline
Dataset & Prediction task & Time unit & \makecell{Window\\ size ($T$)} & \makecell{Minimum\\ segment length ($L_{\min}$)} \\
\hline
HAR      & Classification & Sec. & 96   & 10  \\
ETTm1    & Regression     & Min. & 128  & 13  \\
PTB-XL   & Classification & Sec. & 1000 & 100 \\
S\&P500  & Regression     & Day  & 20   & 4   \\
\hline
\end{tabular*}

\vspace{2pt}
\footnotesize
\noindent{\raggedright \textbf{Note.} Record count ($N$): HAR $N{=}10{,}299$; ETTm1 $N{=}69{,}680$; PTB-XL $N{=}21{,}837$; S\&P500 $N{=}5{,}004$ (daily trading data from 2005-01-01 to 2024-12-31).\par}
\end{table}

\textbf{Predictive Backbone.}
The main experiments use a fixed BiLSTM predictor across the four benchmark datasets to isolate the effect of explainer design. BiLSTM is used as a representative recurrent backbone applicable to both classification and regression tasks. Only the output head is changed according to the task type, with classification heads for HAR and PTB-XL and regression heads for ETTm1 and S\&P500. The trained BiLSTM models achieved accuracies of $0.874$ on HAR and $0.757$ on PTB-XL, with RMSE values of $5.480$ for ETTm1 and $0.014$ for S\&P500.

\textbf{Baseline Explainers.}
GS-SHAP is compared with KernelSHAP \cite{RN13}, TimeSHAP \cite{RN12}, SequenceSHAP \cite{RN10}, WindowSHAP \cite{RN11}, and TSHAP \cite{RN35}. All methods use the same predictive model, samples, background set, masking rule, and perturbation budget whenever applicable. These baselines define players at cell, time-step, subsequence, fixed-window, or sliding-window granularity, whereas GS-SHAP uses dependency-based feature groups with group-specific temporal segments. ShaTS is excluded because it requires expert-defined feature groups.

\textbf{Evaluation Protocols.}
Since explainers produce attributions at different granularities, all outputs are projected onto a common cell-level importance map at the input resolution $(T \times D)$. Each player's attribution is uniformly distributed over its covered cells, and deletion evaluations mask the same fraction of cells per step across methods. Unless stated otherwise, perturbations use mean replacement with feature-wise background means. For methods with different native approximation schemes, we retain their original player definitions and use a common sampled model-query budget for fair comparison under the same prediction-query cost. The same random seed, input samples, and background set are used across methods, with implementation settings summarized in Appendix~\ref{app:impl_details}.

\subsection{Faithfulness Evaluation}

\begin{figure}[h]
  \centering
  \begin{tabular}{cc}
    \includegraphics[width=0.48\linewidth]{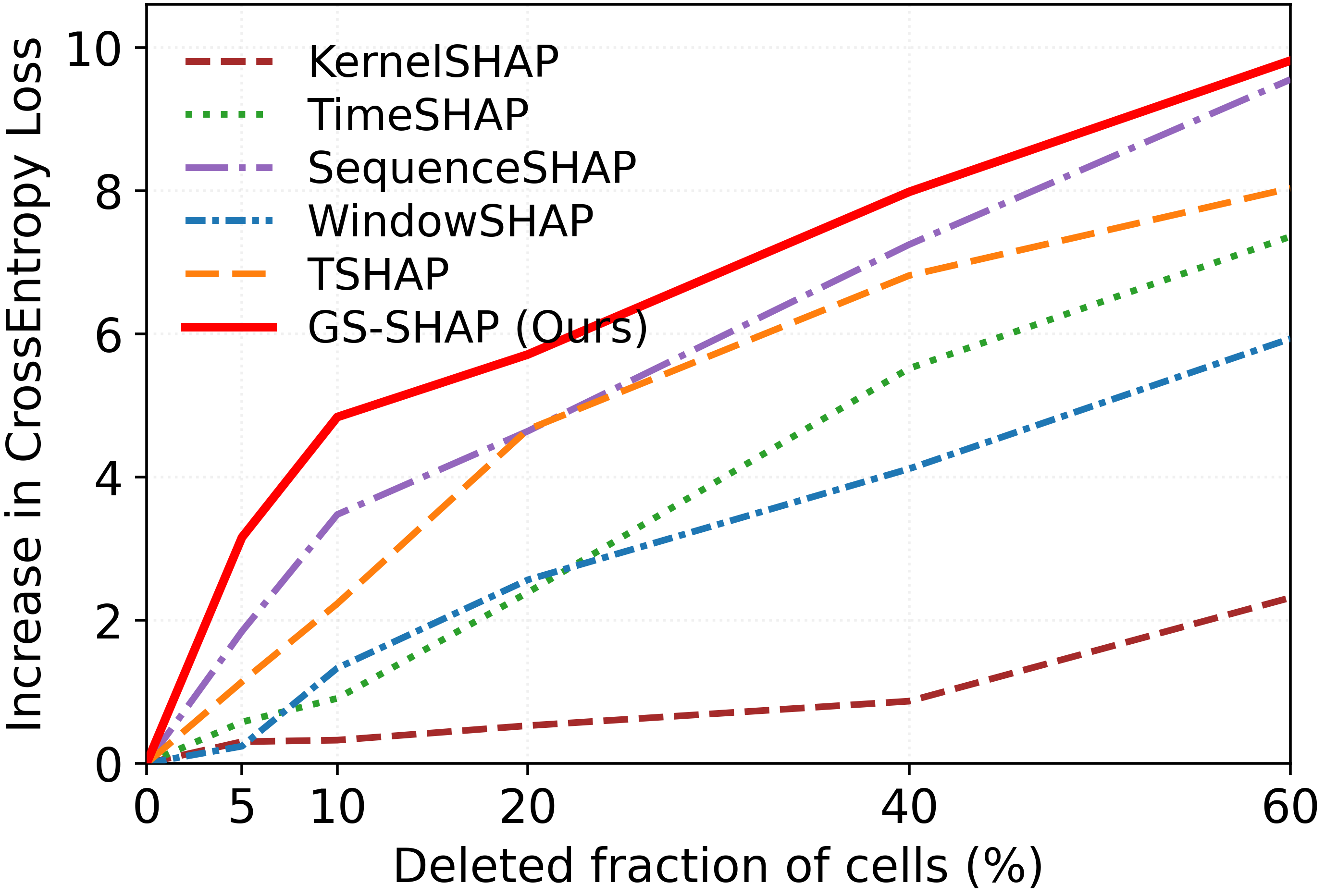} &
    \includegraphics[width=0.48\linewidth]{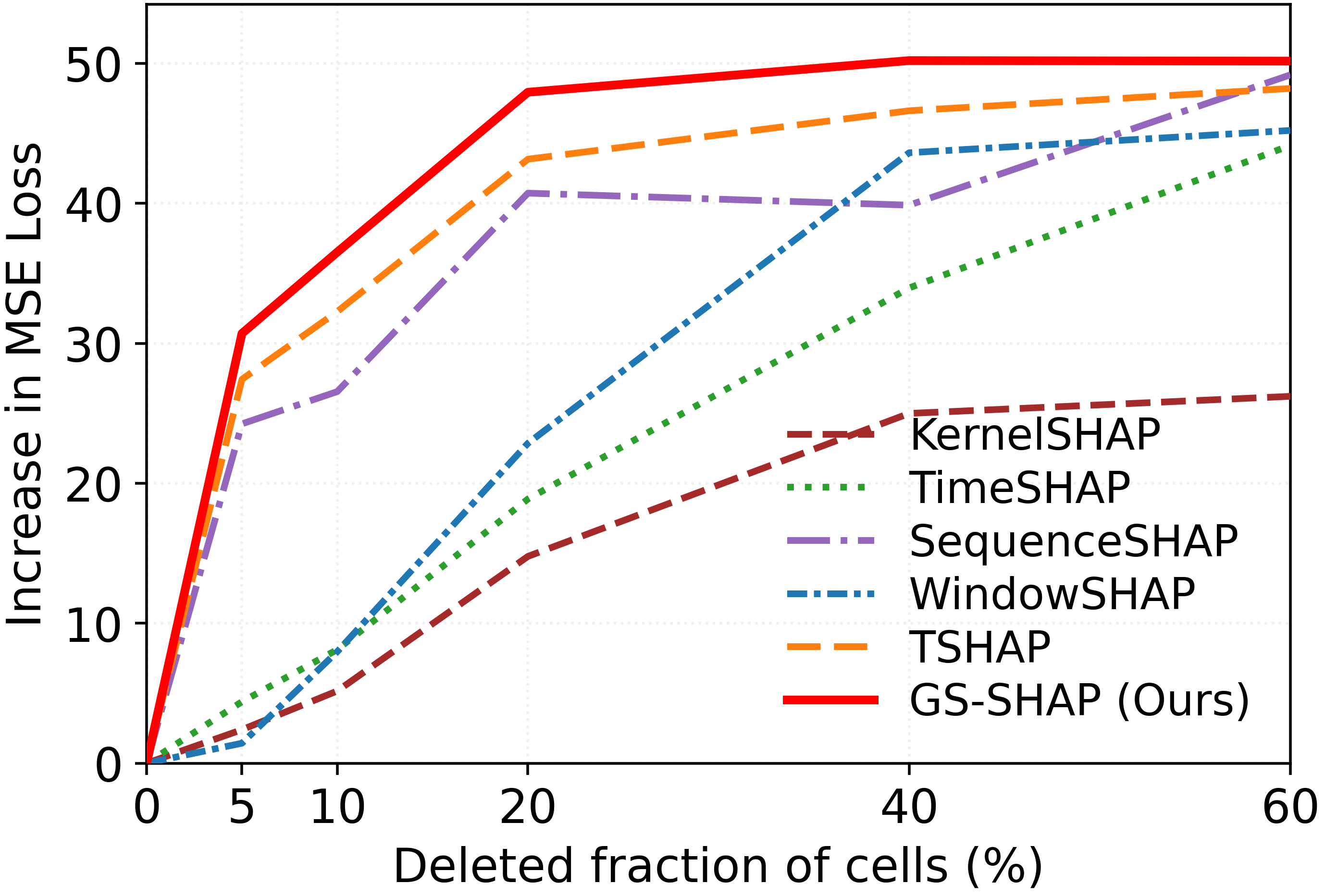} \\
    (a) HAR & (b) ETTm1 \\
    \includegraphics[width=0.48\linewidth]{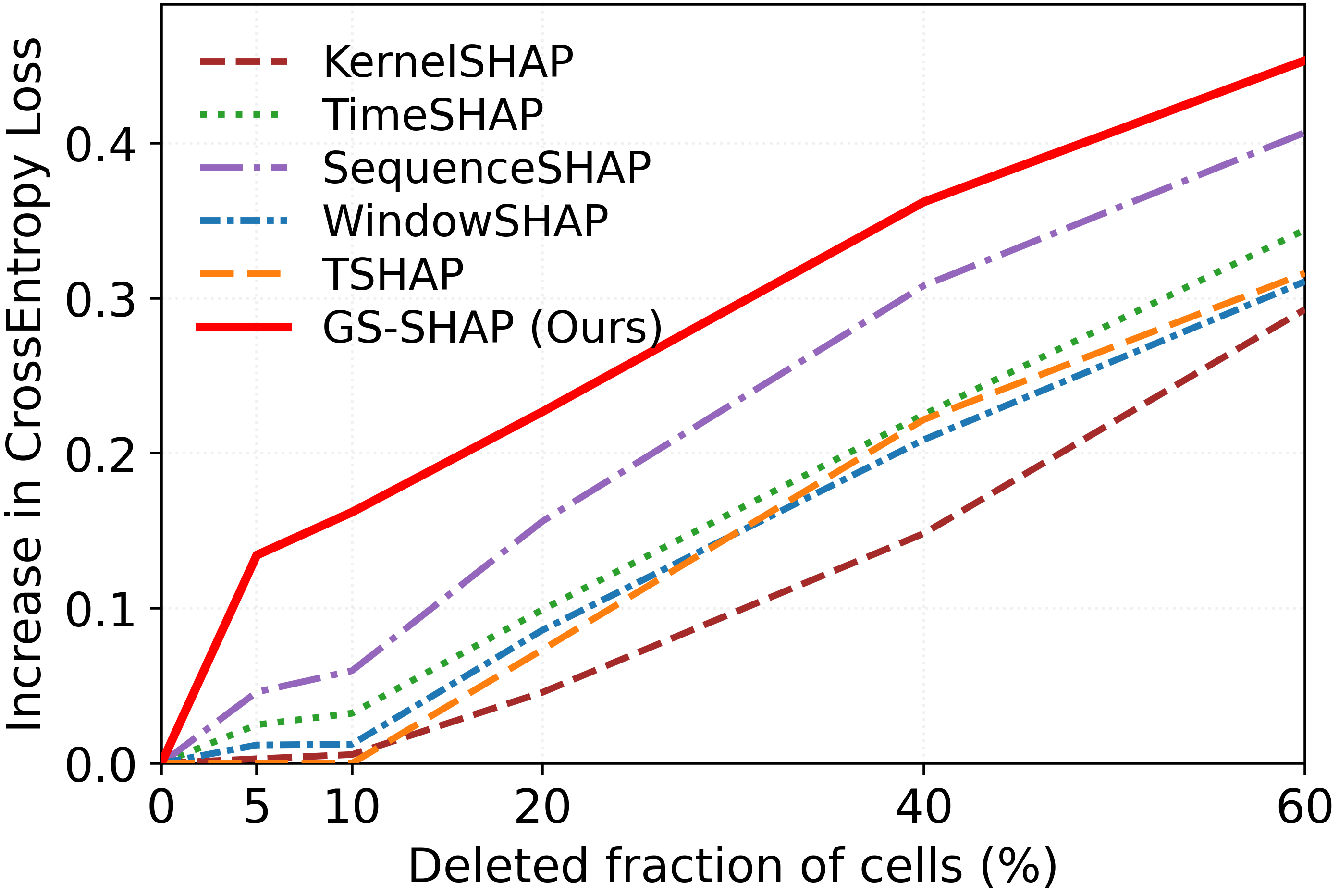} &
    \includegraphics[width=0.48\linewidth]{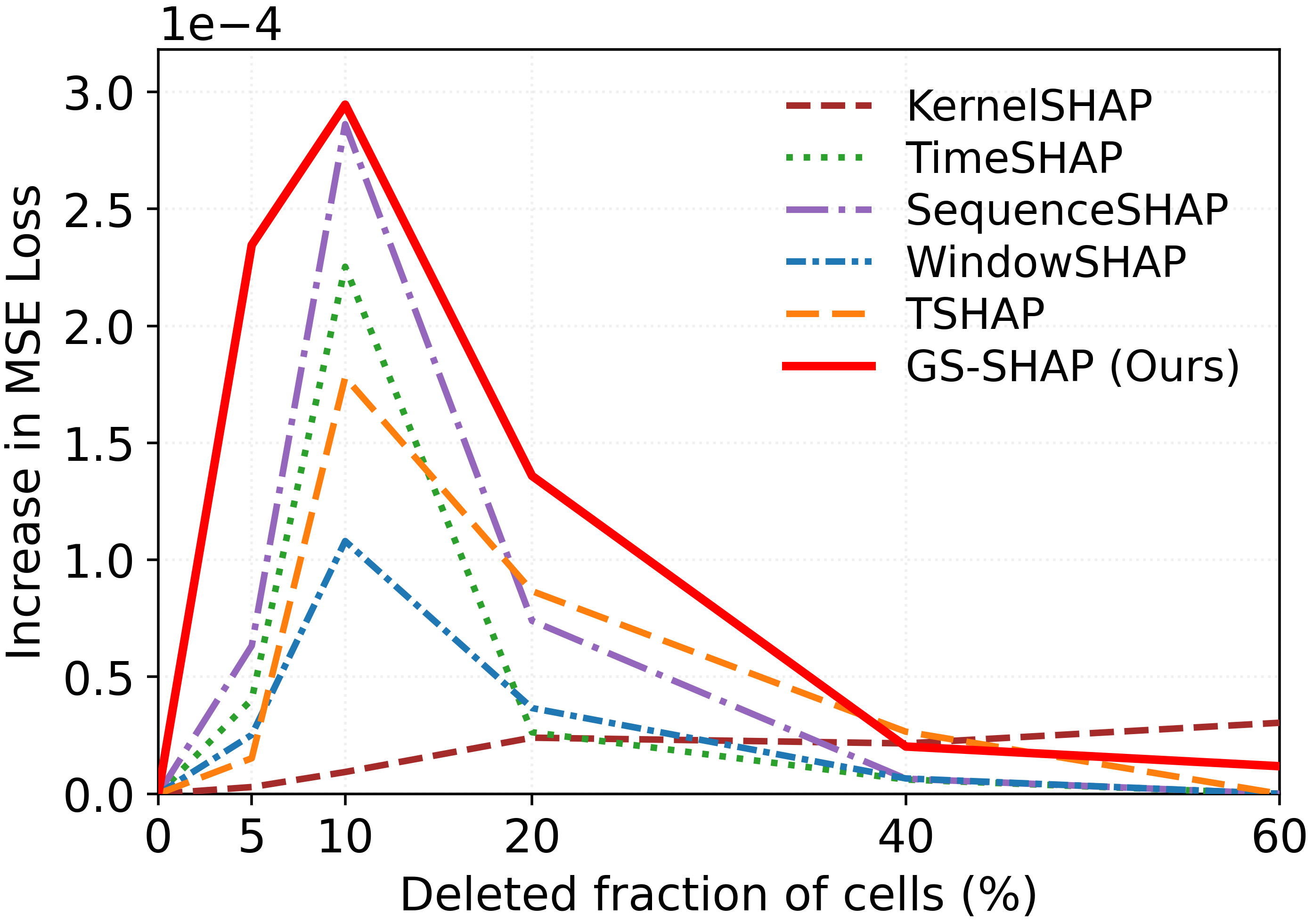} \\
    (c) PTB-XL & (d) S\&P500 \\
  \end{tabular}
  \caption{Deletion-loss curves on four benchmark datasets.}
  \label{fig:del_curves}
\end{figure}

Faithfulness is evaluated using a deletion protocol that progressively masks high-attribution cells and measures the resulting increase in prediction loss. $\Delta$AUC is computed as the area under the deletion-loss increase curve after subtracting the loss at 0\% deletion; higher values indicate stronger degradation after removing high-attribution cells. Because loss scales differ across datasets, $\Delta$AUC is interpreted through within-dataset comparisons across explainers. Figure~\ref{fig:del_curves} shows the deletion curves on four benchmark datasets.

At each deletion step, the highest-importance cells are masked using mean replacement under the common cell-level evaluation protocol described in Section~4.1. Across all datasets, GS-SHAP yields the largest loss increases in Figure~\ref{fig:del_curves} and the highest deletion-based $\Delta$AUC in Table~\ref{tab:delta_auc_main}, indicating better alignment between group-segment players and the multivariate-temporal structures used by the predictor. Sample-wise one-sided paired $t$-tests against the strongest baseline further confirm statistically significant gains across all four datasets.

\begin{table}[h]
\centering
\caption{Deletion $\Delta$AUC on four benchmark datasets.}
\label{tab:delta_auc_main}
\setlength{\tabcolsep}{3pt}
\renewcommand{\arraystretch}{1.05}
\footnotesize
\begin{tabularx}{\linewidth}{l *{4}{>{\centering\arraybackslash}Y}}
\hline
Method & HAR & ETTm1 & PTB-XL & S\&P500 \\
\hline
KernelSHAP   & 0.523  & 10.344 & 0.066 & $1.18{\times}10^{-5}$ \\
TimeSHAP     & 2.294  & 14.867 & 0.098 & $2.41{\times}10^{-5}$ \\
SequenceSHAP & 3.453  & 22.208 & 0.132 & $3.70{\times}10^{-5}$ \\
WindowSHAP   & 1.912  & 17.344 & 0.087 & $1.62{\times}10^{-5}$ \\
TSHAP        & 3.089  & 24.411 & 0.087 & $3.24{\times}10^{-5}$ \\
GS-SHAP      & \textbf{3.955}$^{*}$ & \textbf{26.524}$^{*}$ & \textbf{0.171}$^{*}$ & $\mathbf{5.94{\times}10^{-5}}^{*}$ \\
\hline
\end{tabularx}
\vspace{2pt}
\footnotesize
\noindent{\raggedright \textbf{Note.} Higher is better. $^{*}$ indicates a significant gain over the strongest baseline under a sample-wise one-sided paired $t$-test ($p<0.05$).\par}
\end{table}

These faithfulness gains remain consistent beyond the main recurrent backbone, as confirmed under Transformer and Mamba backbones (Appendix~\ref{app:backbone_generalization}).

Because deletion-based evaluation does not provide ground-truth attribution labels, we further evaluate recovery on synthetic multivariate time series with controlled ground-truth structures, following prior time-series interpretability benchmarks~\cite{RN16}. We consider Single and Two settings with one and two target-generating group-segment players, respectively, and a Distractor setting with correlated but target-irrelevant patterns. Recovery is measured by Cell IoU and Player IoU, with detailed settings provided in Appendix~\ref{app:synthetic_settings}.

\begin{table}[h]
\centering
\caption{Synthetic ground-truth recovery measured by Cell IoU and average Player IoU.}
\label{tab:synthetic_recovery}
\footnotesize
\setlength{\tabcolsep}{3pt}
\renewcommand{\arraystretch}{1.05}
\begin{tabular*}{\linewidth}{@{\extracolsep{\fill}}lccccc}
\hline
Method & \makecell{Single\\Cell IoU} & \makecell{Two\\Cell IoU} & \makecell{Distr.\\Cell IoU} & \makecell{Avg.\\Cell IoU} & \makecell{Avg.\\Player IoU} \\
\hline
KernelSHAP   & 0.075 & 0.077 & 0.089 & 0.080 & 0.293 \\
TimeSHAP     & 0.073 & 0.089 & 0.082 & 0.081 & 0.109 \\
SequenceSHAP & 0.358 & 0.264 & 0.350 & 0.324 & 0.503 \\
WindowSHAP   & 0.080 & 0.093 & 0.083 & 0.085 & 0.115 \\
TSHAP         & 0.080 & 0.093 & 0.083 & 0.085 & 0.118 \\
GS-SHAP      & \textbf{0.447} & \textbf{0.311} & \textbf{0.451} & \textbf{0.403} & \textbf{0.538} \\
\hline
\end{tabular*}
\vspace{2pt}
\noindent{\raggedright \textbf{Note.} Bold indicates the best performance.\par}
\end{table}

As shown in Table~\ref{tab:synthetic_recovery}, GS-SHAP achieves the highest Cell IoU in all three settings and improves the average Cell IoU by $24.4\%$ over the second-best method, SequenceSHAP. It also obtains the highest average Player IoU, demonstrating stronger structural recovery under these controlled settings. Because the synthetic ground-truth patterns are constructed as group-segment structures, this evaluation may favor GS-SHAP and should be interpreted as controlled evidence of structural recovery rather than an unbiased comparison across all attribution structures.

\subsection{Feature Grouping Analysis}

GS-SHAP constructs group-level players to capture cross-variable dependence in multivariate-temporal explanations. To evaluate whether HSIC grouping provides meaningful explanation units, we compare it with Pearson-correlation grouping, no grouping (feature-wise players), and random grouping with the same number of groups as HSIC, while keeping all other settings fixed.
Appendix~\ref{app:supplementary_analysis} provides detailed grouping-comparison results and further supports the role of HSIC grouping as a faithful explanation-unit construction method.
Table~\ref{tab:hsic_groups} reports the HSIC-based groups obtained for the four main benchmark datasets.

\begin{table}[h]
\centering
\caption{HSIC-based feature groups for four datasets.}
\label{tab:hsic_groups}
\setlength{\tabcolsep}{4pt}
\renewcommand{\arraystretch}{1.05}
\footnotesize
\begin{tabularx}{\linewidth}{c YYYY}
\hline
Group & HAR & ETTm1 & PTB-XL & S\&P500 \\
\hline
G0 & TotalAcc-Z & DayCos & I & High \\
G1 & BodyAcc-X, BodyAcc-Z & OilTemp, LoadS & aVL & Gold \\
G2 & BodyAcc-Y, Gyro-Z & Load, OilRate & V1 & Volume, VIX \\
G3 & Gyro-X, Gyro-Y & EnvTemp, EnvRate & V6 & SMA10, SMA20 \\
G4 & TotalAcc-X, TotalAcc-Y & -- & II, aVR & DXY, WTI \\
G5 & -- & -- & III, aVF & Open, Low, Close \\
G6 & -- & -- & V2, V3 & -- \\
G7 & -- & -- & V4, V5 & -- \\
\hline
\end{tabularx}
\vspace{2pt}
\noindent{\raggedright \textbf{Note.} DayCos: cosine time-of-day encoding; LoadS: scaled load. In S\&P500, OHLC denote daily prices and SMA10/20 are computed from close.\par}
\normalsize
\end{table}

\begin{figure}[h]
  \centering
  \begin{tabular}{cc}
    \includegraphics[width=0.48\linewidth]{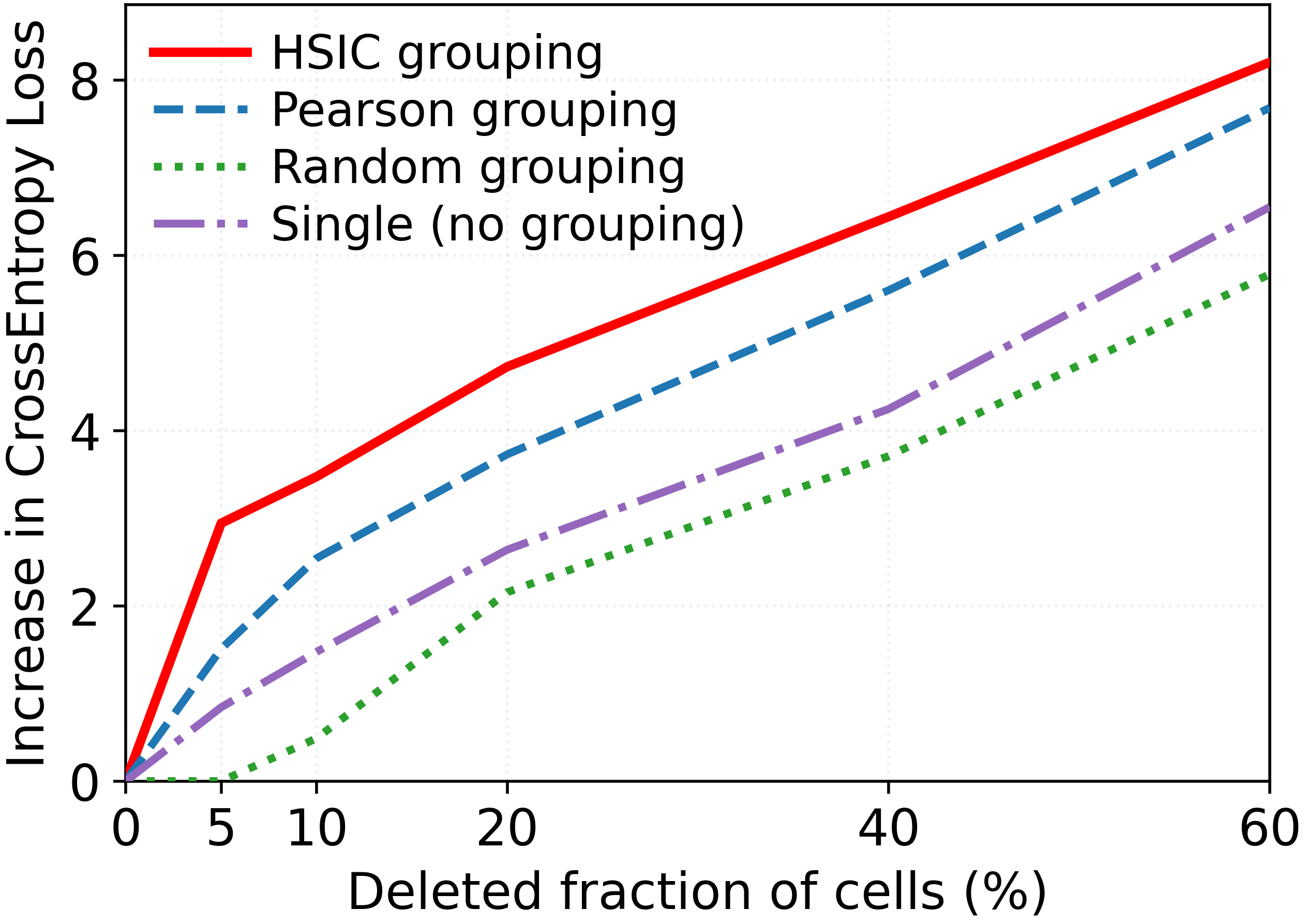} &
    \includegraphics[width=0.45\linewidth]{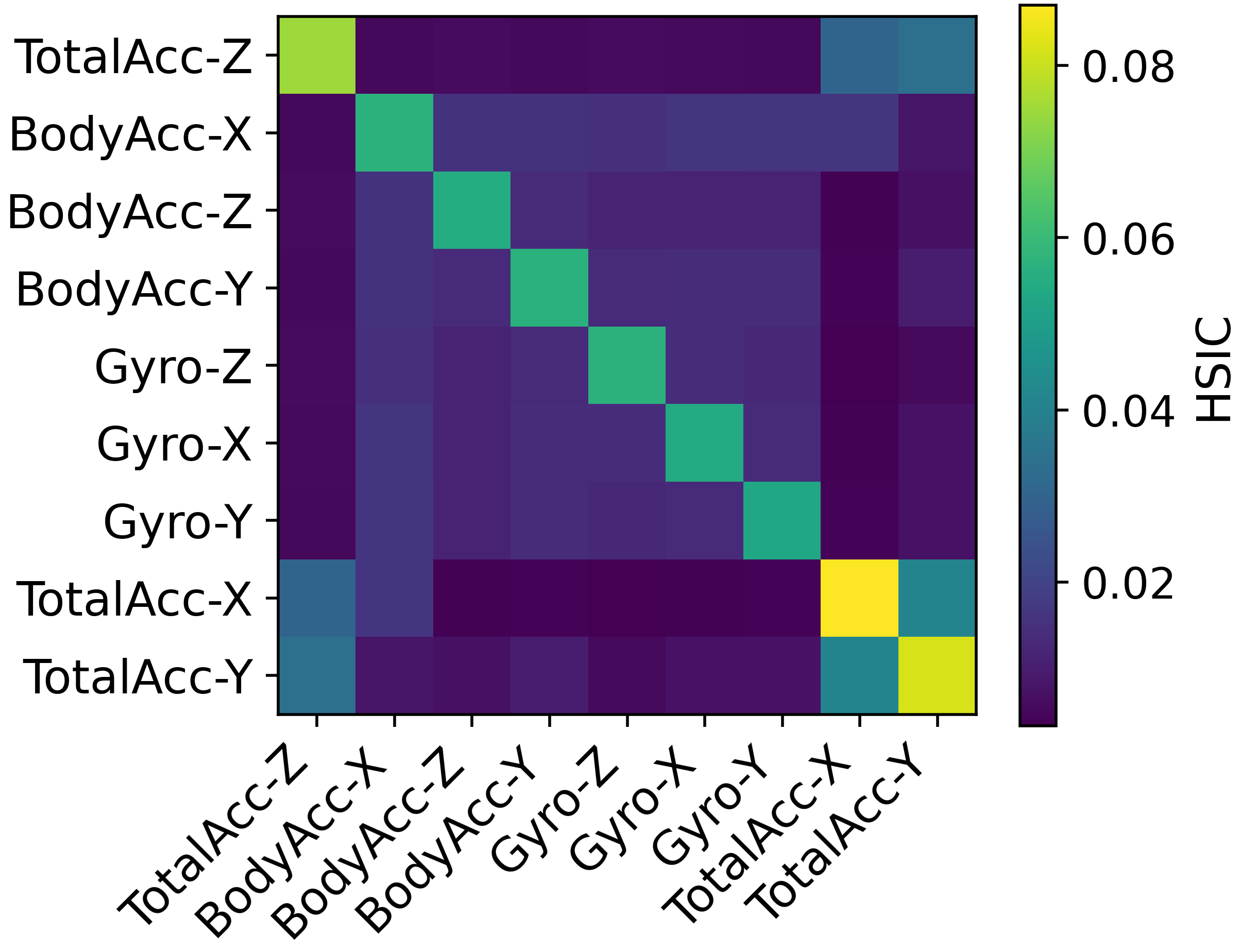} \\
    (a) HAR deletion curve & (b) HAR HSIC similarity matrix \\
    \includegraphics[width=0.48\linewidth]{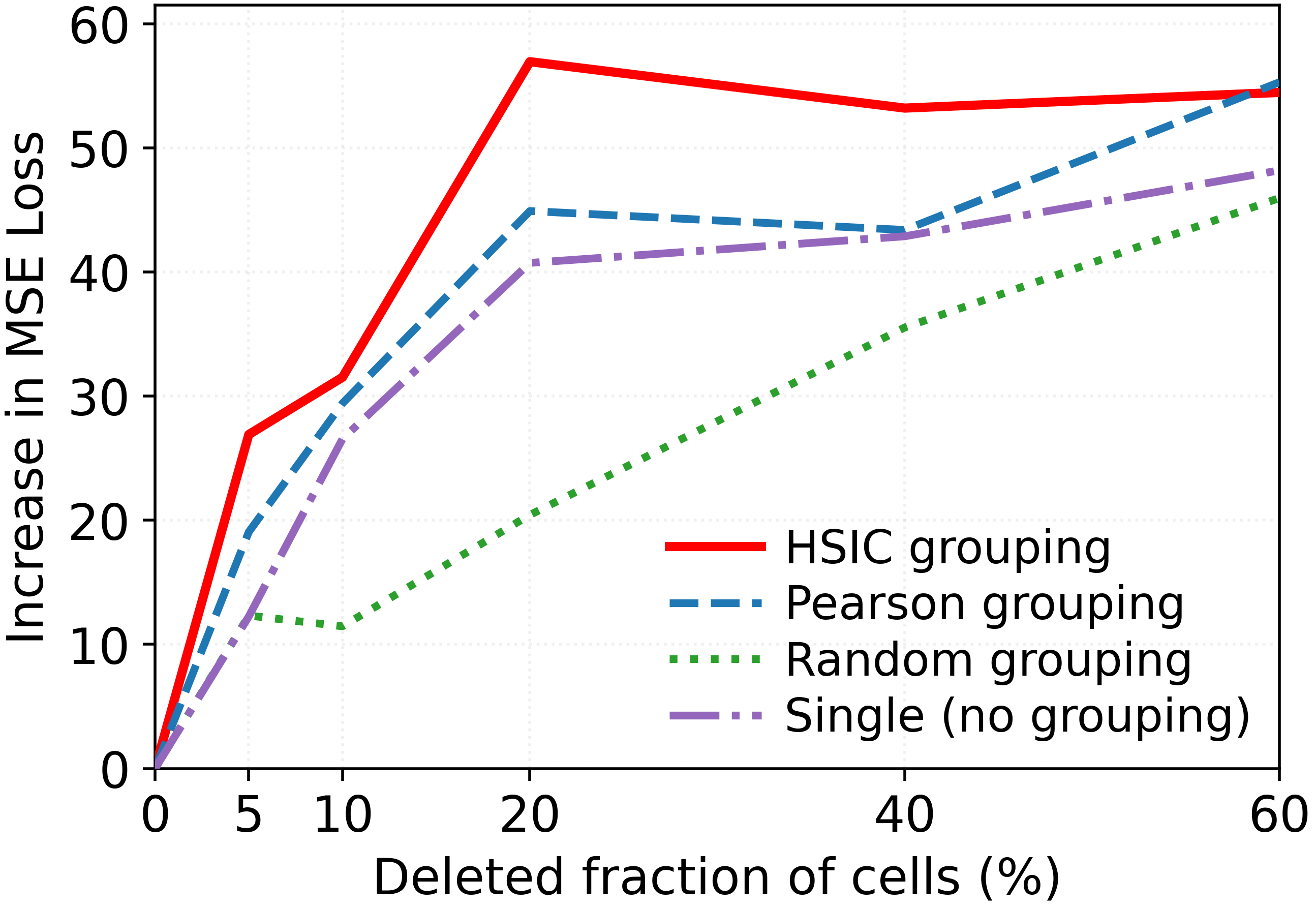} &
    \includegraphics[width=0.45\linewidth]{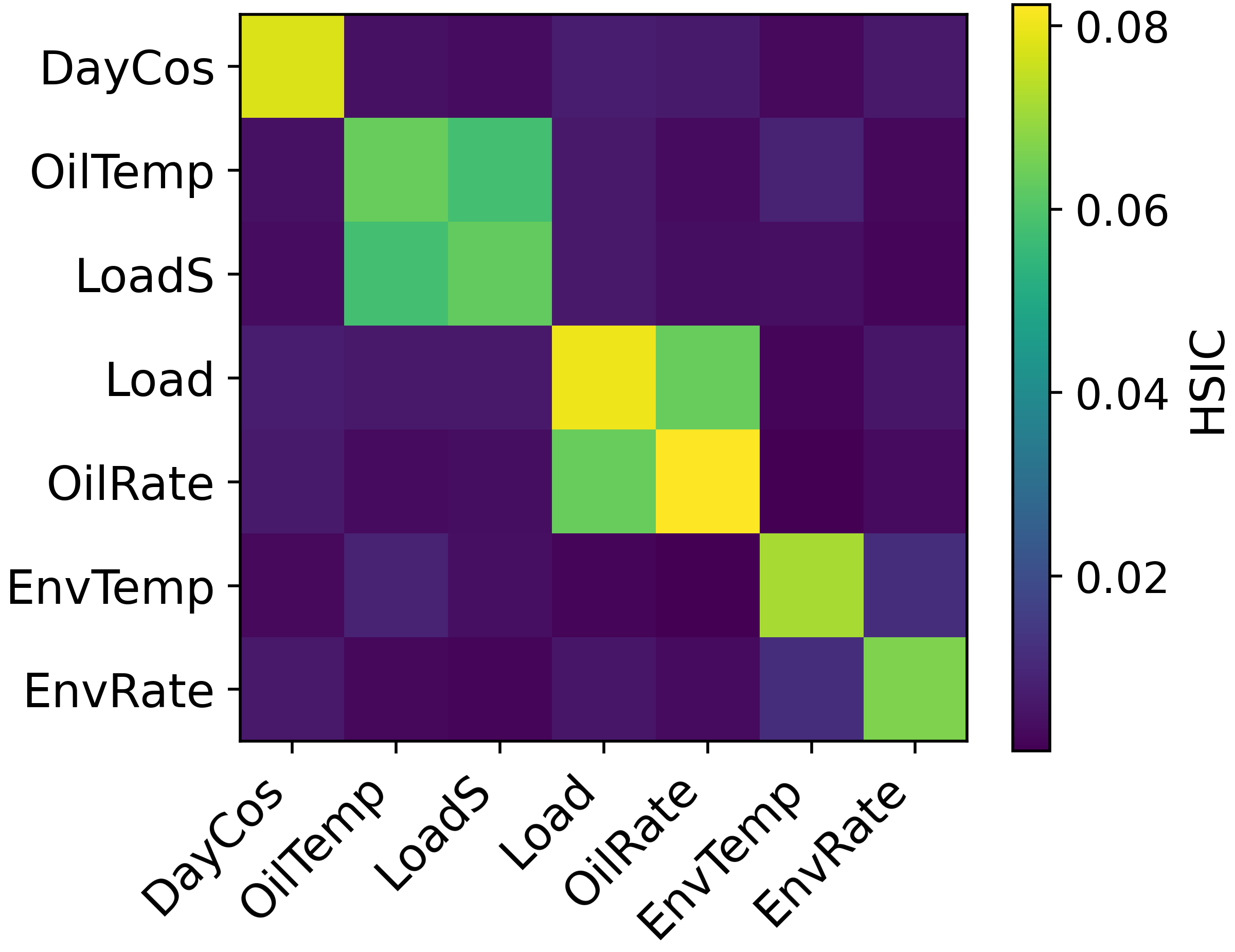} \\
    (c) ETTm1 deletion curve & (d) ETTm1 HSIC similarity matrix \\
  \end{tabular}
  \caption{Comparison of feature grouping strategies on HAR and ETTm1.}
  \label{fig:grouping_har_ettm1}
\end{figure}

The resulting groups are broadly consistent with domain-specific structures: HAR reflects sensor-modality and axis-level relations, ETTm1 separates operation-related and environmental variables, PTB-XL reflects limb and precordial ECG lead structures, and S\&P500 forms groups related to price, volatility, moving averages, and exogenous indicators. Figure~\ref{fig:grouping_har_ettm1} compares grouping strategies and visualizes the reordered HSIC affinity matrices for HAR and ETTm1.

HSIC grouping yields larger loss increases than alternative grouping strategies on HAR and ETTm1. As shown in Appendix~\ref{app:supplementary_analysis}, the deletion-based faithfulness results in Table~\ref{tab:app_delta_auc_grouping} show that HSIC grouping consistently outperforms Pearson, random, and single-feature grouping across all four datasets. Figure~\ref{fig:app_grouping_ptb_sp} further confirms this pattern for PTB-XL and S\&P500, where the reordered HSIC matrices exhibit clear within-group dependence blocks.

To further isolate the contribution of each structural component, Table~\ref{tab:component_ablation} reports a component-level ablation on ETTm1. The full GS-SHAP configuration with HSIC grouping and MMD segmentation achieves the highest deletion-based $\Delta$AUC. Removing HSIC grouping substantially reduces faithfulness, indicating that dependency-based feature groups are important for preserving cross-variable structure. Replacing MMD segmentation with count-matched fixed windows also reduces performance, suggesting that distribution-shift-based temporal segmentation provides more faithful temporal units than fixed intervals. 

\begin{table}[h]
\centering
\caption{HSIC--MMD ablation on ETTm1.}
\label{tab:component_ablation}
\setlength{\tabcolsep}{3pt}
\renewcommand{\arraystretch}{1.05}
\scriptsize
\begin{tabularx}{\linewidth}{l c >{\centering\arraybackslash}X c c}
\hline
Variant & Grouping & Segmentation & $\Delta$AUC $\uparrow$ & Avg. players \\
\hline
GS-SHAP & HSIC & MMD & \textbf{25.86} & 23.84 \\
w/o HSIC & Singleton & MMD & 18.15 & 41.41 \\
w/o MMD & HSIC & Count-matched fixed & 23.25 & 23.84 \\
w/o HSIC,MMD & Singleton & Count-matched fixed & 20.79 & 41.41 \\
\hline
\end{tabularx}
\end{table}
However, the above results still assume a fixed HSIC grouping. To examine the limitation of this assumption, we conducted a controlled synthetic dependency-shift analysis in which the target-relevant feature group changes between the early and late segments. Specifically, both the target-generating group and cross-variable dependence structure change between the early and late segments. This creates time-varying cross-variable dependencies that cannot be perfectly represented by a single time-invariant grouping.

To evaluate whether segment-wise regrouping can better capture such dependency shifts, we compare two variants. \textit{GS-SHAP-Static} estimates HSIC groups once from the training set and reuses them for all test samples, while \textit{GS-SHAP-Dynamic} recomputes HSIC groups within each temporal segment of each explained sample. All other settings are fixed to isolate the effect of static versus segment-wise dynamic grouping. Table~\ref{tab:static_dynamic_synthetic} reports recovery quality and runtime.

\begin{table}[h]
\centering
\caption{Static and Dynamic grouping under synthetic dependency shifts.}
\label{tab:static_dynamic_synthetic}
\footnotesize
\setlength{\tabcolsep}{4pt}
\renewcommand{\arraystretch}{1.05}
\begin{tabular*}{\linewidth}{@{\extracolsep{\fill}}lccc}
\hline
Variant & Cell IoU $\uparrow$ & Player IoU $\uparrow$ & Runtime/sample $\downarrow$ \\
\hline
GS-SHAP-Static  & $0.5314 \pm 0.0531$ & $0.5390 \pm 0.0464$ & $\mathbf{0.1331 \pm 0.0003}$ s \\
GS-SHAP-Dynamic & $\mathbf{0.5448 \pm 0.0199}$ & $\mathbf{0.5462 \pm 0.0242}$ & $0.2269 \pm 0.0012$ s \\
\hline
\end{tabular*}
\end{table}

Dynamic grouping slightly improves Cell IoU and Player IoU by $2.5\%$ and $1.3\%$, respectively, but increases runtime by about $1.7\times$. These results indicate that local regrouping can help under explicit dependency shifts, while static HSIC grouping remains a practical default due to its competitive recovery and lower cost.

\subsection{Robustness and Sensitivity Analysis}

\begin{figure}[h]
  \centering
  \begin{tabular}{cc}
    \includegraphics[width=0.48\linewidth]{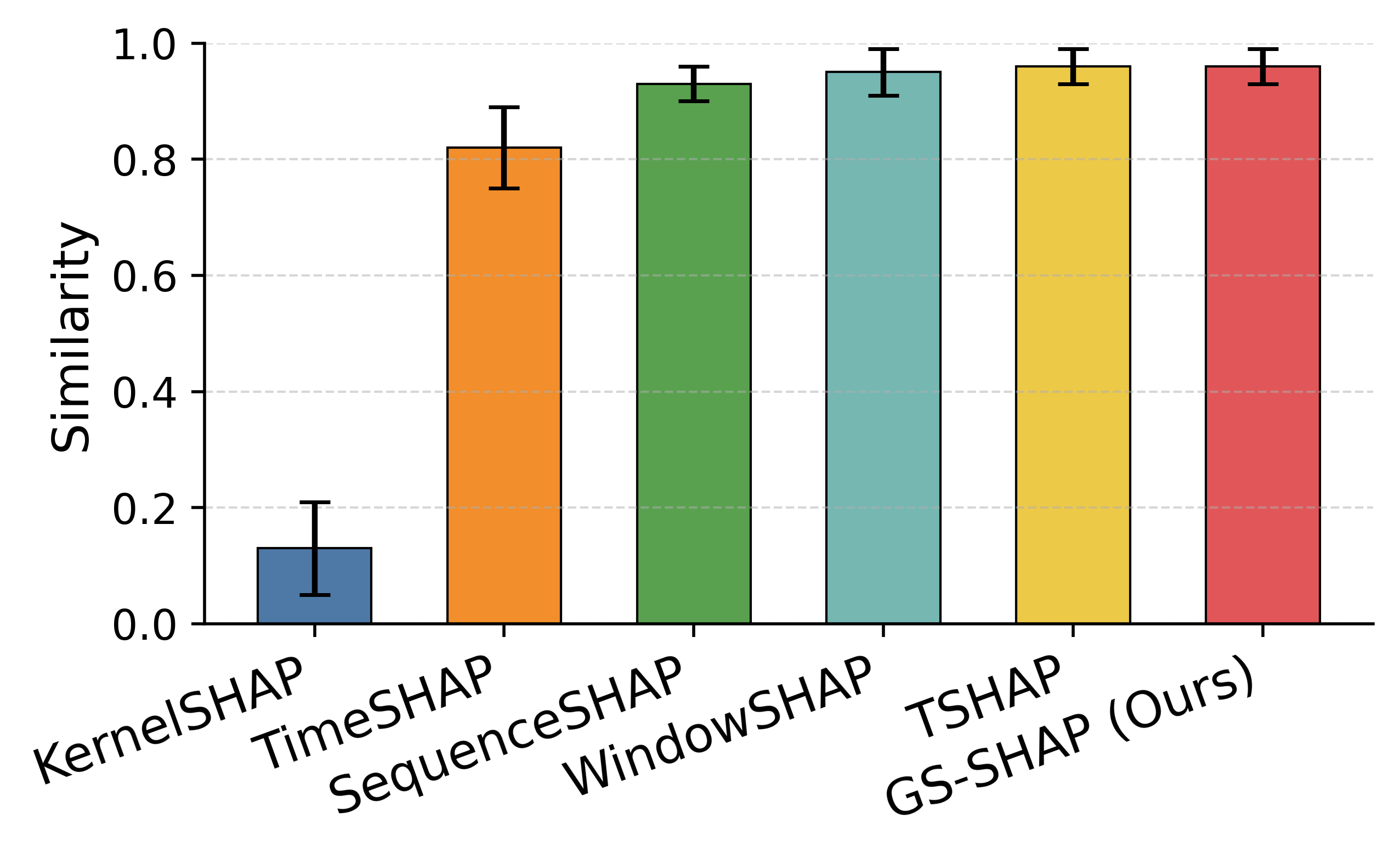} &
    \includegraphics[width=0.48\linewidth]{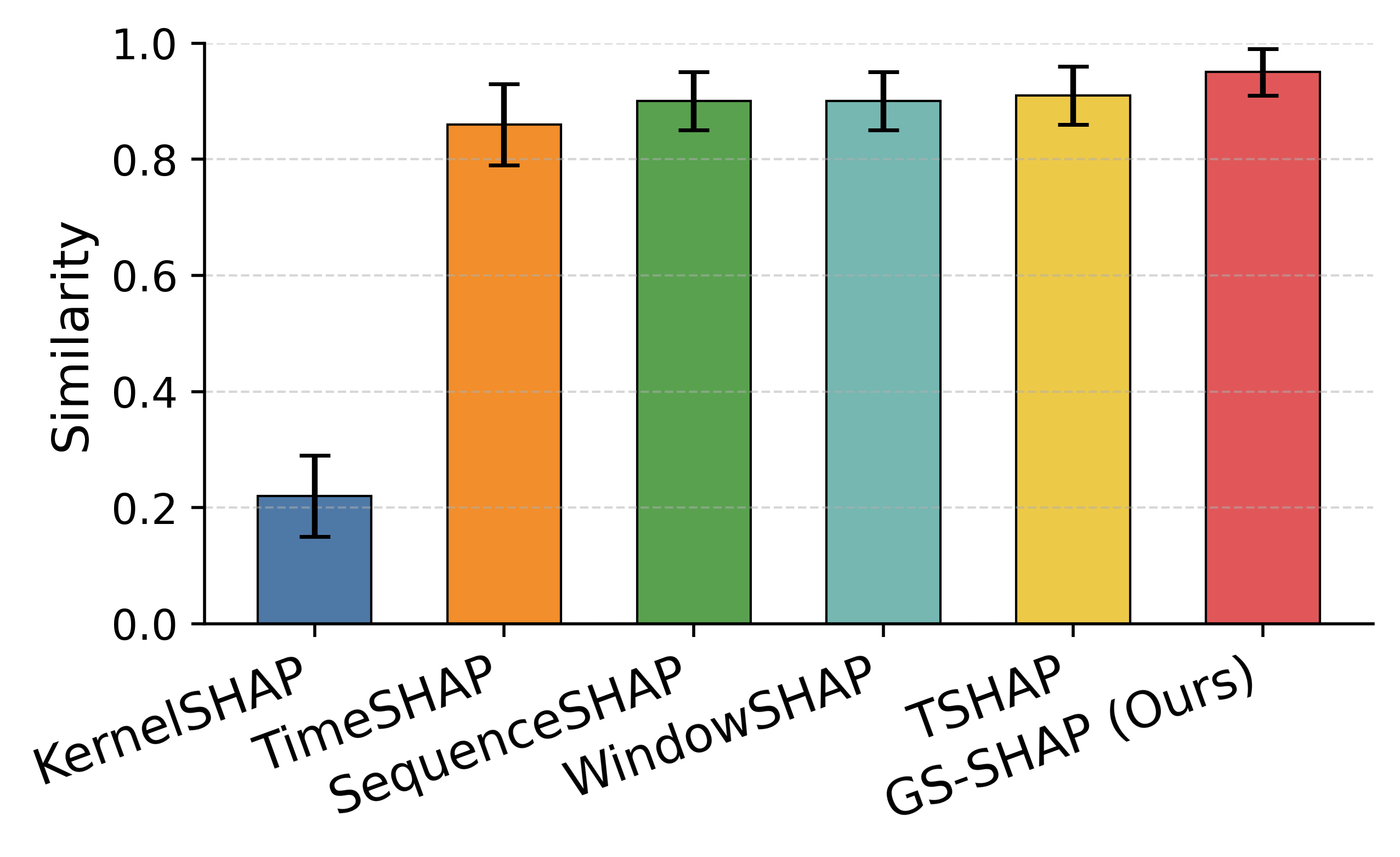} \\
    (a) HAR & (b) ETTm1 \\
    \includegraphics[width=0.48\linewidth]{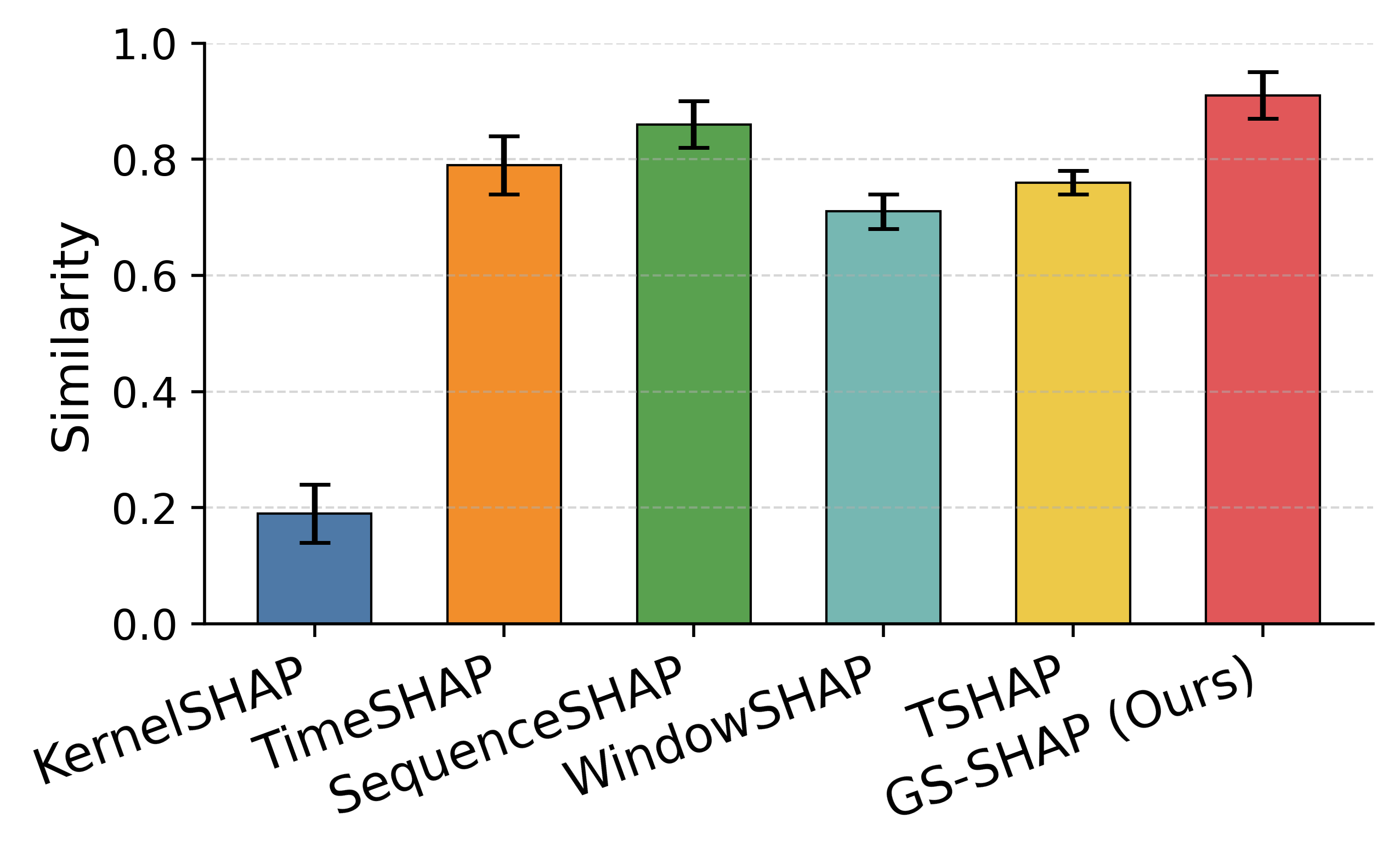} &
    \includegraphics[width=0.48\linewidth]{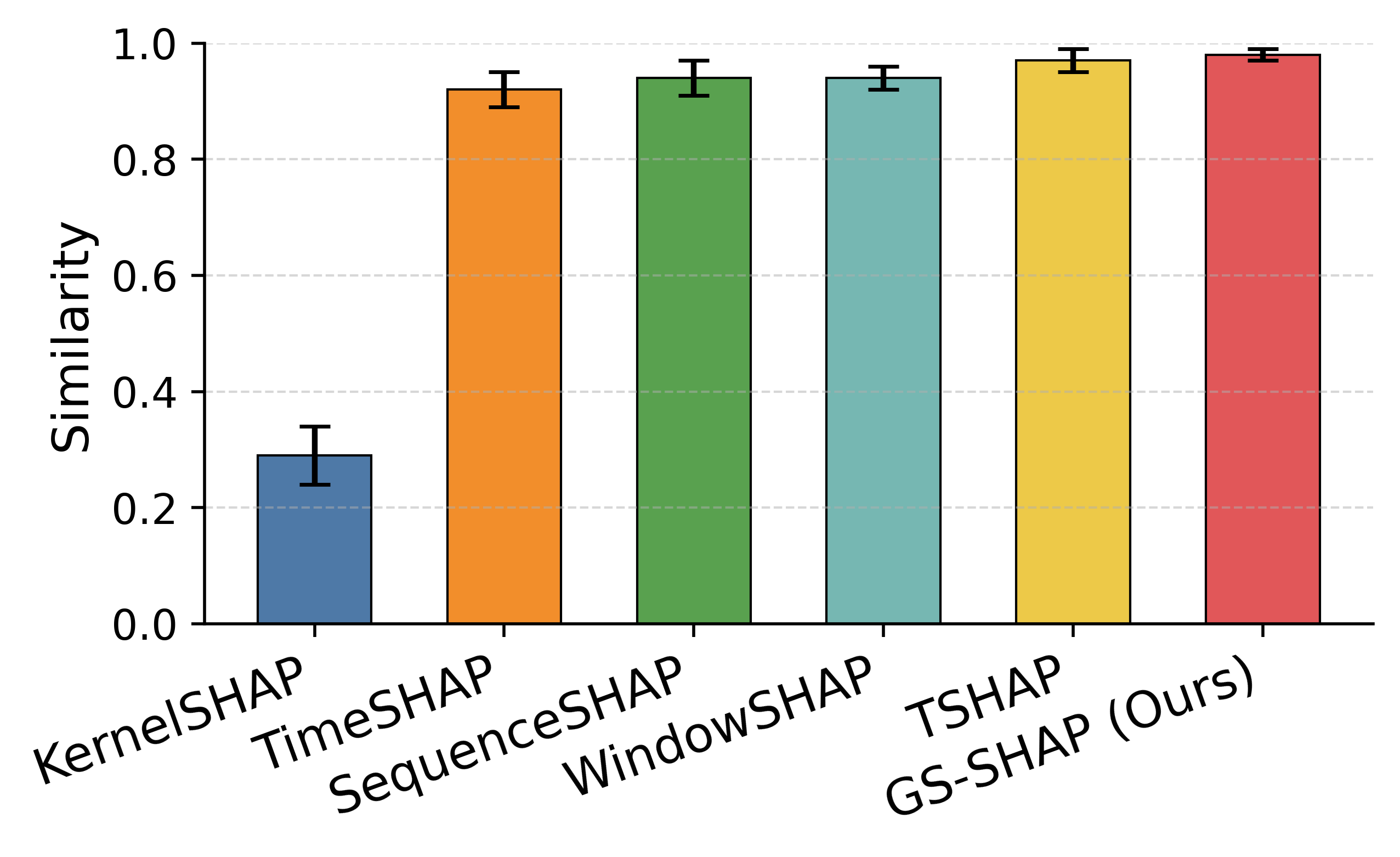} \\
    (c) PTB-XL & (d) S\&P500 \\
  \end{tabular}
  \caption{Attribution stability under background-set changes.}
  \label{fig:background_consistency}
\end{figure}

Time-series explainers can produce different importance patterns under different background sets because Shapley-based masking depends on the reference distribution \cite{RN13,RN29}. For each dataset, we fix one test sample, vary only the background set 10 times, and compute cosine similarity between the resulting cell-level importance maps \cite{RN61,RN62}. Figure~\ref{fig:background_consistency} summarizes the attribution stability under
background-set changes across the four datasets.

GS-SHAP achieves consistently high cosine similarity across the four datasets. WindowSHAP and TSHAP also show relatively high stability, whereas SequenceSHAP and TimeSHAP vary by dataset and KernelSHAP is the most sensitive. This suggests that group-segment players reduce background-induced variation by aggregating attributions over dependency-based feature groups and temporal segments.

We further examine sensitivity on ETTm1, where heterogeneous feature scales can amplify masking-induced shifts. Tables~\ref{tab:minseg_sens} and~\ref{tab:masking_sens} report the effects of varying the minimum segment length $L_{\min}$ and masking baseline. Changes in $L_{\min}$ cause only modest variation in $\Delta$AUC and $\Delta$Loss@0.60, and zero/noise masking reduces $\Delta$AUC by $4.0\%$ and $4.83\%$, respectively. This local ETTm1 sweep, separate from the main benchmark configuration, indicates that GS-SHAP is not highly sensitive to the segmentation constraint or masking baseline.

\begin{table}[h]
\centering
\caption{Sensitivity to minimum segment length.}
\label{tab:minseg_sens}
\setlength{\tabcolsep}{4pt}
\renewcommand{\arraystretch}{1.05}
\footnotesize
\begin{tabularx}{\columnwidth}{>{\centering\arraybackslash}m{0.22\columnwidth} *{2}{>{\centering\arraybackslash}Y}}
\hline
MinSegLen ($L_{\min}$) & $\Delta$AUC & $\Delta$Loss@0.60 \\
\hline
4  & 27.119 & 83.548 \\
6  & 27.219 & 85.276 \\
8  & 27.640 & 86.765 \\
10 & 28.904 & 90.036 \\
12 & 27.900 & 88.410 \\
16 & 26.494 & 84.181 \\
\hline
\end{tabularx}
\end{table}

\vspace{-0.4em}

\begin{table}[h]
\centering
\caption{Sensitivity to masking baseline.}
\label{tab:masking_sens}
\setlength{\tabcolsep}{4pt}
\renewcommand{\arraystretch}{1.05}
\footnotesize
\begin{tabularx}{\columnwidth}{>{\centering\arraybackslash}m{0.22\columnwidth} *{2}{>{\centering\arraybackslash}Y}}
\hline
Masking mode & $\Delta$AUC & $\Delta$Loss@0.60 \\
\hline
Mean  & 28.904 & 90.036 \\
Zero  & 27.747 & 84.071 \\
Noise & 27.507 & 83.641 \\
\hline
\end{tabularx}
\end{table}

\subsection{Computational Efficiency}
\label{sec:computational_efficiency}

We report end-to-end wall-clock runtime in seconds per sample. All methods use the same predictive model, mean-replacement masking, and matched model-query budgets $M\in\{10,20,30,50\}$. Runtime is averaged over 100 test samples and includes player construction, masked-sample generation, and model queries. KernelSHAP is evaluated with the same sampled perturbation budget rather than exhaustive coalition enumeration.

\begin{table}[h]
\centering
\caption{Runtime by approximation budget on ETTm1.}
\label{tab:runtime_budget}
\setlength{\tabcolsep}{3pt}
\renewcommand{\arraystretch}{1.05}
\scriptsize
\begin{tabularx}{\linewidth}{l *{4}{>{\centering\arraybackslash}Y}}
\hline
Method & $M{=}10$ & $M{=}20$ & $M{=}30$ & $M{=}50$ \\
\hline
KernelSHAP   & $0.006 \pm 0.002$ & $0.011 \pm 0.003$ & $0.016 \pm 0.002$ & $0.026 \pm 0.002$ \\
TimeSHAP     & $0.404 \pm 0.005$ & $0.805 \pm 0.011$ & $1.202 \pm 0.013$ & $2.001 \pm 0.030$ \\
SequenceSHAP & $0.232 \pm 0.030$ & $0.365 \pm 0.032$ & $0.498 \pm 0.032$ & $0.764 \pm 0.035$ \\
WindowSHAP   & $0.233 \pm 0.021$ & $0.333 \pm 0.023$ & $0.434 \pm 0.025$ & $0.534 \pm 0.029$ \\
TSHAP        & $0.194 \pm 0.022$ & $0.252 \pm 0.027$ & $0.334 \pm 0.022$ & $0.465 \pm 0.029$ \\
GS-SHAP      & $0.172 \pm 0.031$ & $0.249 \pm 0.029$ & $0.324 \pm 0.030$ & $0.473 \pm 0.030$ \\
\hline
\end{tabularx}
\vspace{2pt}
\footnotesize
\noindent{\raggedright \textbf{Note.} Mean $\pm$ std.\ runtime (seconds per sample) over 100 test samples.\par}
\end{table}

Table~\ref{tab:runtime_budget} shows that GS-SHAP remains within the runtime range of window- and subsequence-based explainers. In power-system forecasting, GS-SHAP is on average $3.38\times$ faster than TimeSHAP across approximation budgets. At $M{=}50$, it takes $0.473$ s per sample, comparable to TSHAP and WindowSHAP.

This efficiency follows from the reduced player space induced by HSIC grouping and group-wise temporal segmentation. Appendix~\ref{app:player_space_reduction} presents an average 96.9\% reduction in the original cell-level player space, and Appendix~\ref{app:runtime_scaling} shows that GS-SHAP generally runs faster than TimeSHAP and SequenceSHAP. These results indicate that group-segment players reduce the computational burden of Shapley attribution while simplifying the explanation space.

\section{Case Study}

To assess the practical interpretability of GS-SHAP, we conduct a case study on the S\&P500 next-day return prediction model using two market regimes. Regimes are defined by the magnitude of the next-day return $r_{t+1}$. The high-volatility regime is defined as $|r_{t+1}|\ge 3.0\%$ \cite{RN64}; we select April 29, 2022, when earnings-related shocks in major technology stocks and macro uncertainty coincided with a $3.63\%$ decline in the S\&P500 \cite{RN66}. The stable regime is defined as $|r_{t+1}|<0.2\%$ \cite{RN63}, and we select a test-period window satisfying this condition.

Both cases explain the model prediction for an input window of $T$ trading days ending at time $t$. Group-segment attributions are computed using the same HSIC-based feature grouping and group-wise MMD-based temporal segmentation.

\begin{figure}[h]
  \centering
  \begin{tabular}{cc}
    \includegraphics[width=0.48\linewidth]{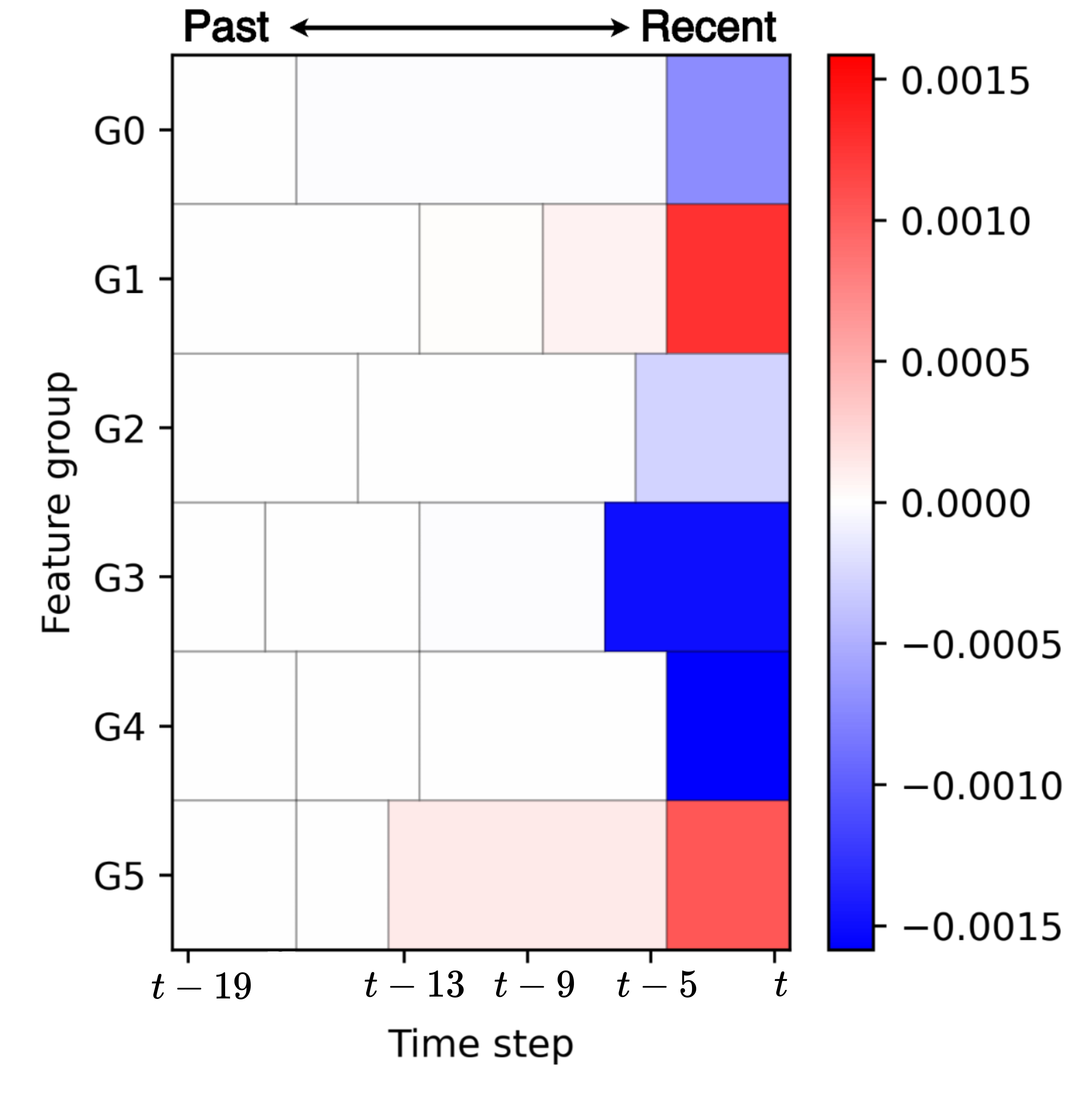} &
    \includegraphics[width=0.48\linewidth]{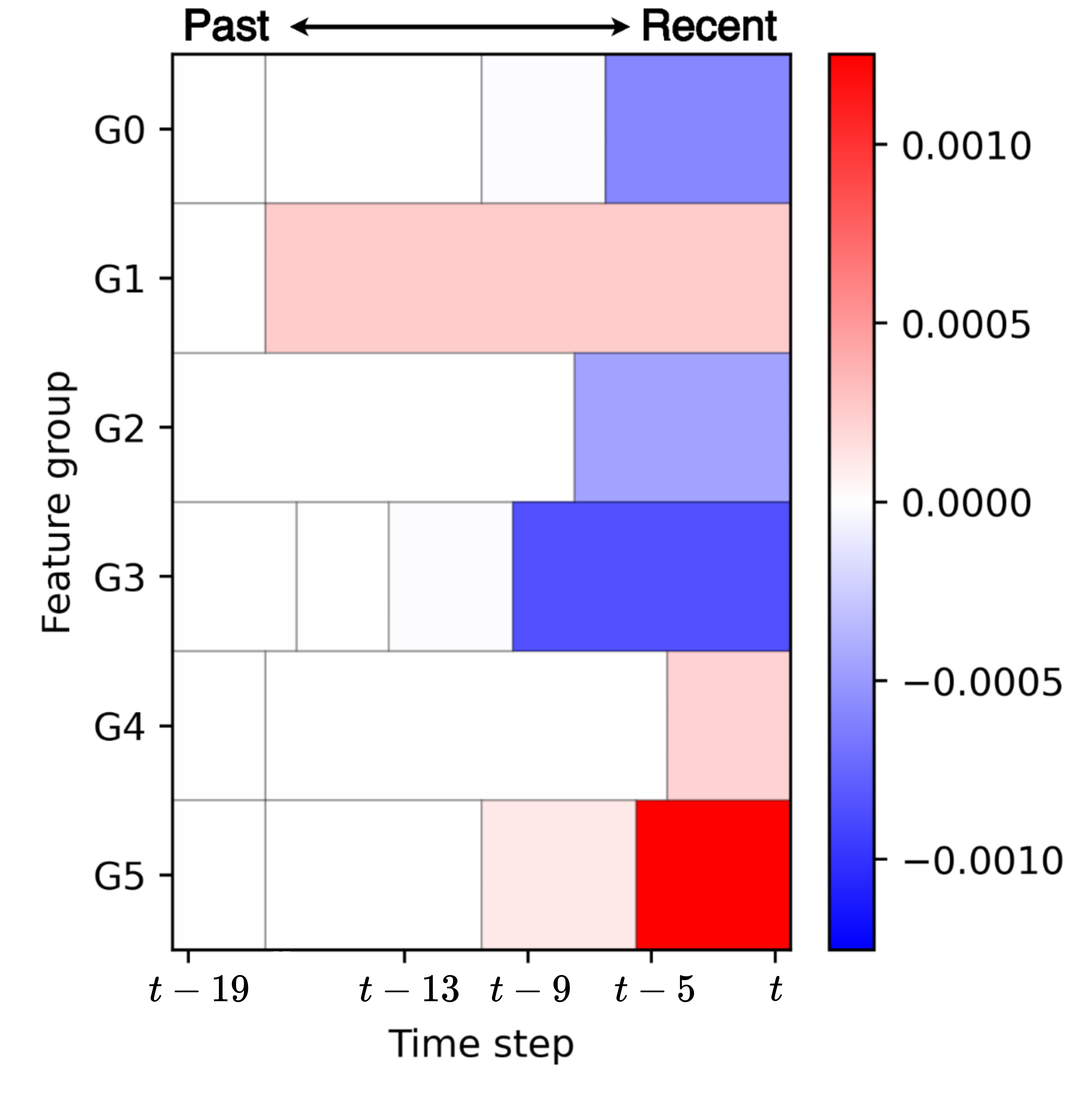} \\
    (a) High-volatility regime & (b) Stable regime \\
  \end{tabular}
  \caption{GS-SHAP attribution maps under high-volatility and stable market regimes.}
  \label{fig:sp500_case}
\end{figure}

Figure~\ref{fig:sp500_case} visualizes the group-segment importance distributions, where each colored block corresponds to a temporal segment within a feature group. Table~\ref{tab:sp500_top_players} reports the top-ranked players. Segment IDs are group-specific because MMD-based segmentation is performed independently for each feature group.

\begin{table}[h]
\centering
\caption{Top group-segment players in the S\&P500 case study.}
\label{tab:sp500_top_players}

\footnotesize
\setlength{\tabcolsep}{2pt}
\renewcommand{\arraystretch}{1.08}

\begin{tabularx}{\linewidth}{%
  >{\centering\arraybackslash}m{0.18\linewidth}
  >{\centering\arraybackslash}m{0.10\linewidth}
  >{\centering\arraybackslash}m{0.15\linewidth}
  >{\centering\arraybackslash}m{0.08\linewidth}
  X
  >{\centering\arraybackslash}m{0.16\linewidth}
  >{\centering\arraybackslash}m{0.05\linewidth}
}
\hline
\makecell[c]{Regime} &
\makecell[c]{Segment\\ID} &
\makecell[c]{Time\\interval} &
\makecell[c]{Group} &
\makecell[c]{Features\\in group} &
\makecell[c]{SHAP\\value ($\phi$)} &
\makecell[c]{Rank} \\
\hline

\multirow{3}{=}{\centering\makecell[c]{High\\Volatility}} &
S4 & $[t-6,t]$  & G3 & SMA10, SMA20       & $-0.009$ & 1 \\
& S4 & $[t-4,t]$ & G4 & DXY, WTI           & $-0.006$ & 2 \\
& S4 & $[t-4,t]$ & G1 & Gold               & $0.005$  & 3 \\
\hline

\multirow{3}{=}{\centering \raisebox{-2.0ex}{Stable}} &
S4 & $[t-9,t]$   & G3 & SMA10, SMA20      & $-0.008$ & 1 \\
& S4 & $[t-5,t]$  & G5 & Open, Low, Close  & $0.006$  & 2 \\
& S2 & $[t-17,t]$ & G1 & Gold              & $0.004$  & 3 \\
\hline
\end{tabularx}

\vspace{2pt}
\noindent{\raggedright \textbf{Note.} Time interval denotes the contiguous segment covered by each group-segment player. Segment IDs are group-specific.\par}
\normalsize
\end{table}

Across both regimes, the model assigns substantial importance to recent temporal segments. In the high-volatility regime, recent segments of $G_3$ and $G_4$ contribute negatively, whereas $G_1$ contributes positively. This suggests that recent moving-average patterns and global risk-related indicators serve as negative predictive cues before the sharp decline, while GS-SHAP localizes these signals to specific feature groups and intervals.

In the stable regime, the recent segment of $G_5$ provides the strongest positive contribution, while the influence of $G_3$ and $G_4$ weakens. This suggests that the model relies more on price-level and short-term dynamics than on exogenous indicators in this example, illustrating a regime-dependent shift in the model's information sources.

\vspace{-0.8em}

\section{Conclusion}

We presented GS-SHAP, a Shapley-based explanation framework for multivariate time-series models that defines players as intersections of dependency-based feature groups and distribution-shift-based temporal segments. By using group-segment players, GS-SHAP preserves coupled multivariate-temporal patterns as coherent attribution units and reduces the fragmentation caused by feature-only or time-only explanations.

Across four real-world domains, GS-SHAP achieved the highest deletion-based faithfulness among the compared SHAP-style time-series explainers on all benchmark datasets. In synthetic evaluations with controlled ground-truth structures, it also achieved 24.4\% higher average Cell IoU than the strongest baseline. Component ablation and additional analyses showed that both HSIC grouping and MMD segmentation contribute to faithful explanations, while GS-SHAP remains robust under background and masking variations and maintains computational efficiency.

The S\&P500 case study further showed that GS-SHAP can reveal regime-dependent feature groups and temporal intervals under high-volatility and stable market conditions. Future work will explore adaptive grouping for strongly regime-dependent dependencies, more efficient Shapley approximation, and broader validation across model architectures and application domains.

\appendix
\makeatletter
\renewcommand{\thetable}{\thesection\arabic{table}}
\renewcommand{\thefigure}{\thesection\arabic{figure}}
\@addtoreset{table}{section}
\@addtoreset{figure}{section}
\makeatother

\section{Theoretical Complexity Analysis}
\label{app:complexity_analysis}
This appendix summarizes the stage-wise theoretical complexity of GS-SHAP.

\begin{table}[h]
\centering
\caption{Stage-wise theoretical complexity of GS-SHAP.}
\label{tab:app_complexity}
\setlength{\tabcolsep}{4pt}
\renewcommand{\arraystretch}{1.08}
\scriptsize
\begin{tabularx}{\linewidth}{l Y}
\hline
Stage & Complexity \\
\hline
HSIC grouping & $O(D^2 n^2)$ one-time preprocessing cost for $n$ sampled observations \\
MMD segmentation & Data-dependent, bounded in practice by $L_{\min}$ and $J_{\max}$ \\
Player construction & $O(|P|)$, where $|P|=\sum_{k=1}^{K}J_k$ \\
Shapley attribution & $O(M|P|)$ model evaluations per explained sample \\
\hline
\end{tabularx}
\end{table}

\section{Implementation Settings}
\label{app:impl_details}

This appendix summarizes the experimental settings.

\begin{table}[h]
\centering
\caption{Predictive backbone settings.}
\label{tab:app_backbone_settings}
\setlength{\tabcolsep}{4pt}
\renewcommand{\arraystretch}{0.95}
\scriptsize
\begin{tabularx}{\linewidth}{l Y}
\hline
Component & Setting \\
\hline
BiLSTM & 2 layers; hidden size 64; dropout 0.2; Adam optimizer \\
Transformer & 2 encoder layers; model dimension 64; 4 attention heads; feed-forward dimension 128; dropout 0.1; Adam optimizer \\
Mamba & 2 Mamba blocks; model dimension 64; state dimension 16; convolution width 4; expansion factor 2; Adam optimizer \\
Training objective & Cross-entropy for classification; mean squared error for regression \\
\hline
\end{tabularx}
\end{table}

\clearpage

\begin{table}[h]
\centering
\caption{GS-SHAP implementation settings.}
\label{tab:app_gsshap_settings}
\setlength{\tabcolsep}{4pt}
\renewcommand{\arraystretch}{0.95}
\scriptsize
\begin{tabularx}{\linewidth}{l Y}
\hline
Component & Setting \\
\hline
Background set & Up to 3000 training samples drawn with a fixed seed; feature-wise means used for masking \\
HSIC sampling & Up to $N_{\mathrm{HSIC}}=3000$ pooled training observations for HSIC affinity estimation \\
HSIC kernel & 1D RBF kernel with median-heuristic bandwidth \\
Grouping & Spectral clustering on HSIC affinity with eigengap-based group-number selection \\
MMD segmentation & Group-wise recursive split search using unbiased MMD with RBF kernel and candidate stride 1 \\
Split threshold & Permutation-calibrated threshold at significance level $\alpha{=}0.05$ using 50 permutations \\
Stopping rule & Stop if interval length $<2L_{\min}$, no significant split is found, or dataset-specific maximum segment count is reached \\
Shapley budget & Same model-query budget $M$ across methods \\
\hline
\end{tabularx}
\end{table}

\section{Backbone Generalization}
\label{app:backbone_generalization}
This appendix reports results under Transformer and Mamba backbones.

\begin{table}[H]
\centering
\caption{Transformer-backbone $\Delta$AUC.}
\label{tab:app_transformer_backbone}
\setlength{\tabcolsep}{4pt}
\renewcommand{\arraystretch}{0.95}
\scriptsize
\begin{tabularx}{\linewidth}{l *{4}{>{\centering\arraybackslash}Y}}
\hline
Method & HAR & ETTm1 & PTB-XL & S\&P500 \\
\hline
KernelSHAP   & 0.1200 & 12.5000 & 0.4300 & $2.20{\times}10^{-6}$ \\
TimeSHAP     & 0.8342 & 34.1171 & 1.1917 & $8.60{\times}10^{-6}$ \\
SequenceSHAP & 2.8697 & 51.2423 & 1.2809 & $1.53{\times}10^{-4}$ \\
WindowSHAP   & 0.7903 & 33.7690 & 0.5223 & $3.55{\times}10^{-6}$ \\
TSHAP        & 1.5050 & 70.5619 & 0.5238 & $8.60{\times}10^{-6}$ \\
GS-SHAP      & \textbf{3.3251} & \textbf{73.4596} & \textbf{1.5683} & $\mathbf{4.15{\times}10^{-4}}$ \\
\hline
\end{tabularx}
\end{table}

\begin{table}[H]
\centering
\caption{Mamba-backbone $\Delta$AUC.}
\label{tab:app_mamba_backbone}
\setlength{\tabcolsep}{4pt}
\renewcommand{\arraystretch}{0.95}
\scriptsize
\begin{tabularx}{\linewidth}{l *{4}{>{\centering\arraybackslash}Y}}
\hline
Method & HAR & ETTm1 & PTB-XL & S\&P500 \\
\hline
KernelSHAP   & 0.3100 & 18.2000 & 0.5200 & $3.00{\times}10^{-4}$ \\
TimeSHAP     & 2.4689 & 39.0052 & 1.1606 & $4.25{\times}10^{-4}$ \\
SequenceSHAP & 3.5578 & 52.1768 & 1.2566 & $5.11{\times}10^{-3}$ \\
WindowSHAP   & 2.4441 & 38.8553 & 0.5863 & $1.90{\times}10^{-4}$ \\
TSHAP        & 3.3617 & 58.5995 & 0.6084 & $1.90{\times}10^{-4}$ \\
GS-SHAP      & \textbf{3.7926} & \textbf{64.6709} & \textbf{1.2867} & $\mathbf{6.42{\times}10^{-3}}$ \\
\hline
\end{tabularx}
\end{table}

\section{Synthetic Data Settings}
\label{app:synthetic_settings}
This appendix summarizes the synthetic-data settings used for ground-truth recovery.

\begin{table}[H]
\centering
\caption{Synthetic data settings.}
\label{tab:app_synthetic_settings}
\setlength{\tabcolsep}{4pt}
\renewcommand{\arraystretch}{0.95}
\scriptsize
\begin{tabularx}{\linewidth}{l Y}
\hline
Component & Setting \\
\hline
Input size & Window size 100; 12 variables \\
Ground-truth structure & 4 variable groups and 5 temporal segments \\
Single setting & One target-generating group-segment player determines the target \\
Two setting & Two target-generating group-segment players jointly determine the target \\
Distractor setting & Correlated but target-irrelevant group-segment patterns are added \\
Noise & Additive Gaussian noise in target-irrelevant regions \\
Evaluation & Cell IoU for each setting and average Player IoU \\
\hline
\end{tabularx}
\end{table}

\section{Supplementary Grouping Results}
\label{app:supplementary_analysis}
This appendix reports additional results for the feature-grouping analysis.

\begin{table}[H]
\centering
\caption{Grouping-strategy $\Delta$AUC.}
\label{tab:app_delta_auc_grouping}
\setlength{\tabcolsep}{3pt}
\renewcommand{\arraystretch}{0.95}
\scriptsize
\begin{tabularx}{\linewidth}{l *{4}{>{\centering\arraybackslash}Y}}
\hline
Grouping method & HAR & ETTm1 & PTB-XL & S\&P500 \\
\hline
HSIC grouping        & \textbf{3.225} & \textbf{28.339} & \textbf{0.128} & $\mathbf{1.11{\times}10^{-4}}$ \\
Pearson grouping     & 2.714 & 24.103 & 0.119 & $8.30{\times}10^{-5}$ \\
Random grouping      & 1.680 & 16.241 & 0.087 & $5.78{\times}10^{-5}$ \\
Single (no grouping) & 2.053 & 22.108 & 0.110 & $7.69{\times}10^{-5}$ \\
\hline
\end{tabularx}
\end{table}

\begin{figure}[H]
  \centering
  \begin{tabular}{cc}
    \includegraphics[width=0.40\linewidth]{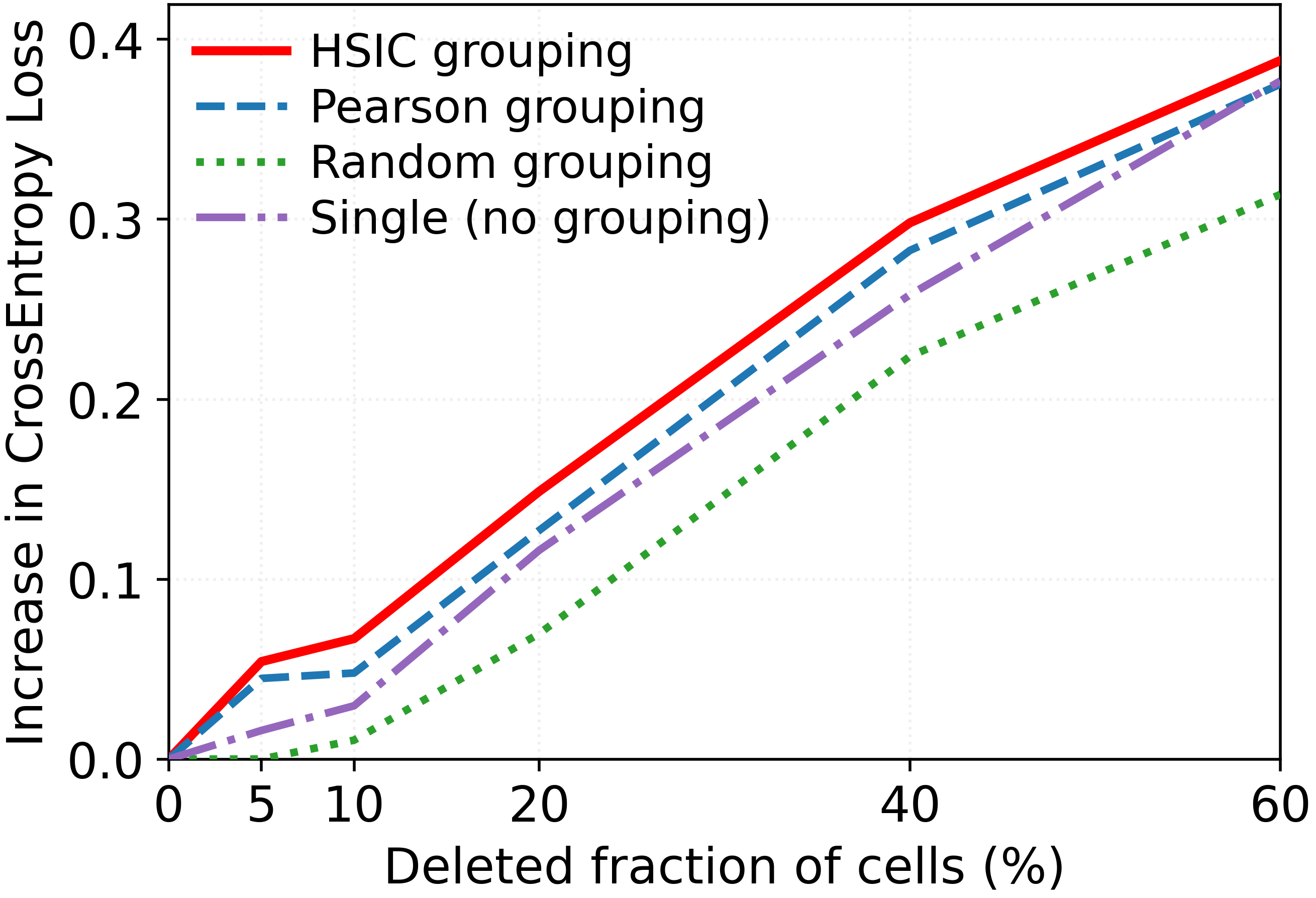} &
    \includegraphics[width=0.40\linewidth]{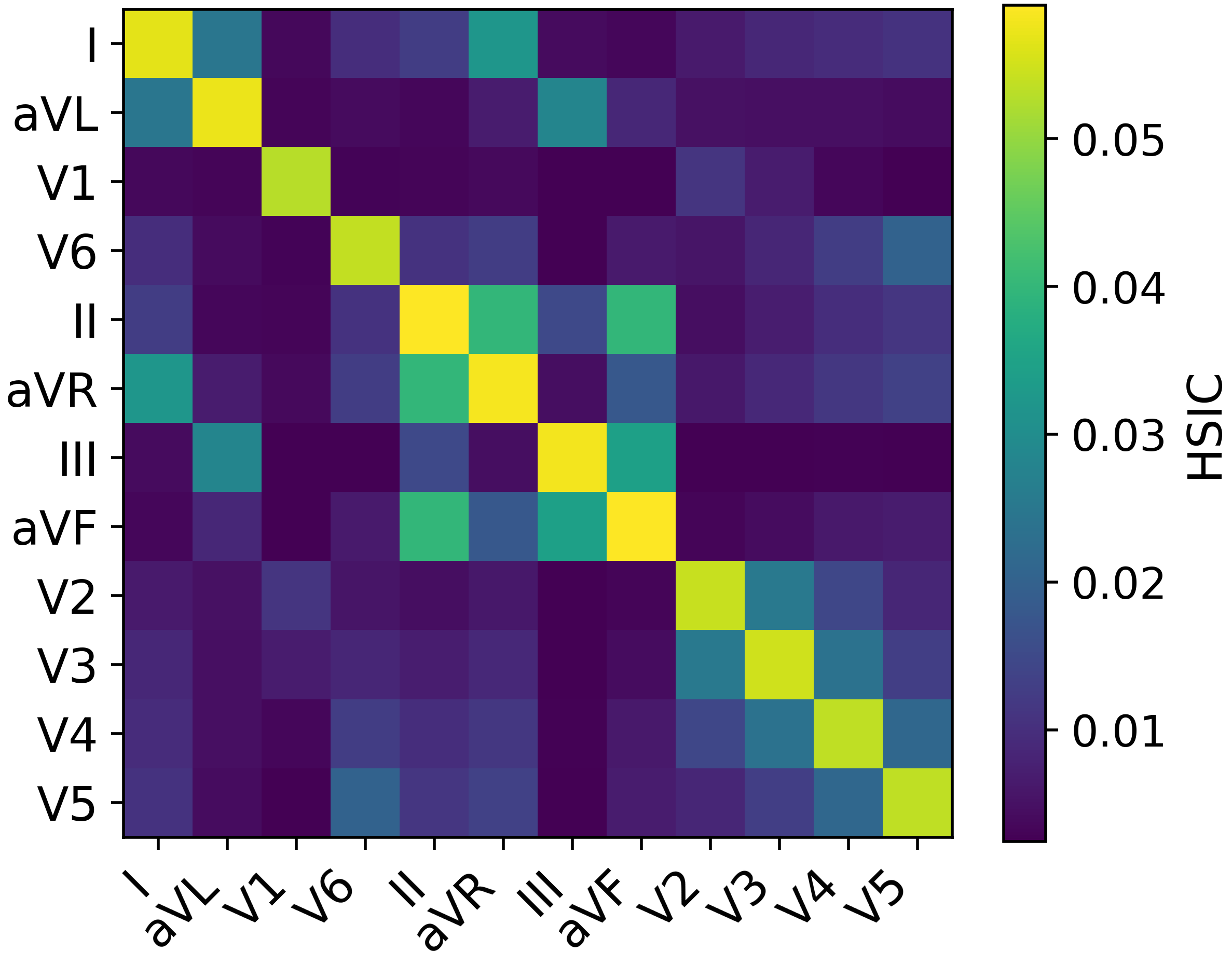} \\
    (a) PTB-XL deletion curve & (b) PTB-XL HSIC similarity matrix \\
    \includegraphics[width=0.40\linewidth]{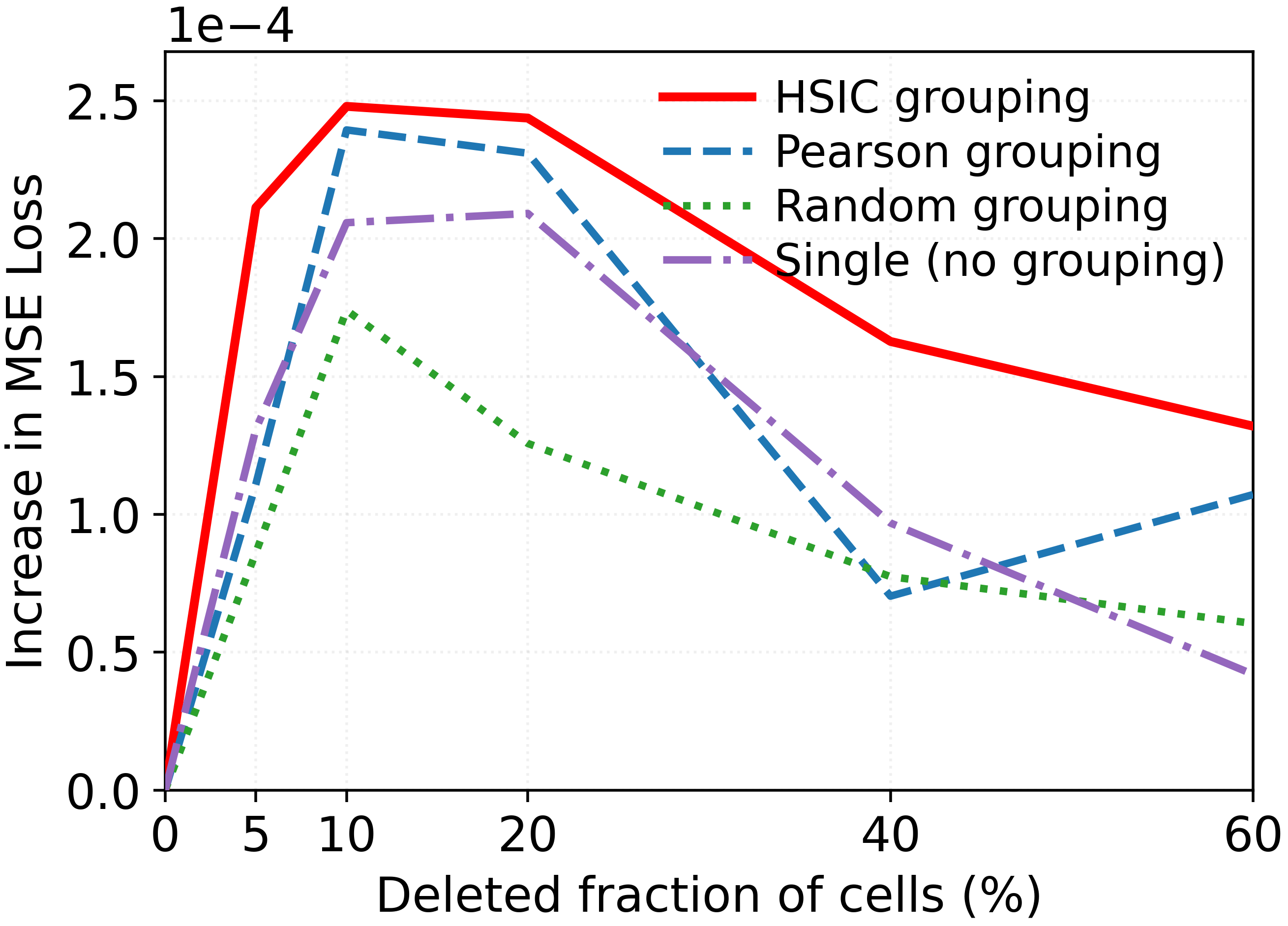} &
    \includegraphics[width=0.40\linewidth]{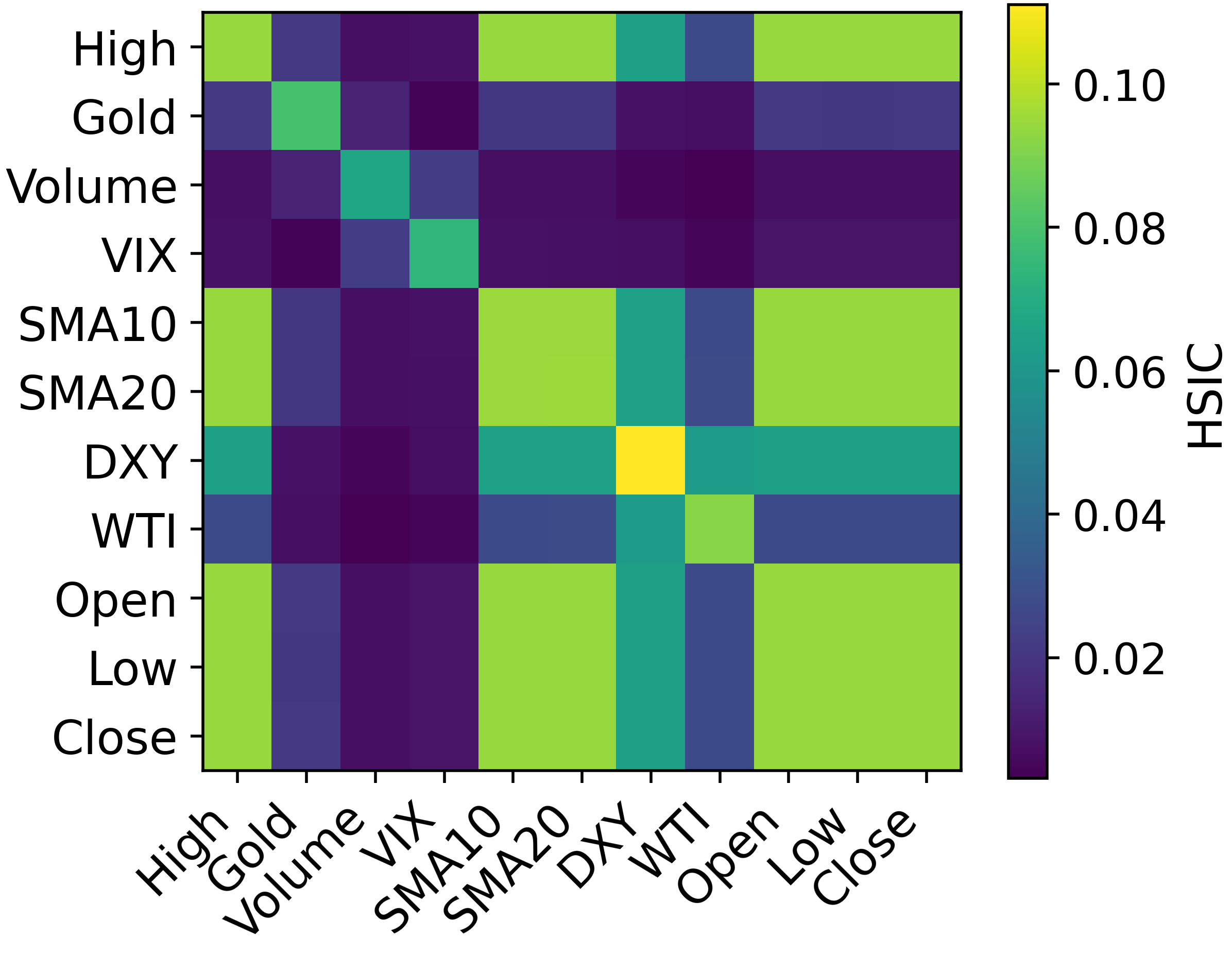} \\
    (c) S\&P500 deletion curve & (d) S\&P500 HSIC similarity matrix \\
  \end{tabular}
  \caption{Additional grouping-strategy results on PTB-XL and S\&P500.}
  \label{fig:app_grouping_ptb_sp}
\end{figure}

\section{Player-Space Reduction}
\label{app:player_space_reduction}
This appendix summarizes the reduction in the GS-SHAP player space.

\begin{table}[H]
\centering
\caption{Player-space reduction.}
\label{tab:app_player_reduction}
\setlength{\tabcolsep}{4pt}
\renewcommand{\arraystretch}{0.95}
\scriptsize
\begin{tabularx}{\linewidth}{l *{3}{>{\centering\arraybackslash}Y}}
\hline
Dataset & Original cells & Avg. players & Reduction \\
\hline
HAR     & 1,152  & 35.76 & 96.90\% \\
ETTm1   & 672    & 23.96 & 96.43\% \\
PTB-XL  & 12,000 & 31.80 & 99.74\% \\
S\&P500 & 220    & 11.91 & 94.59\% \\
\hline
\end{tabularx}
\vspace{2pt}
\footnotesize
\noindent{\raggedright \textbf{Note.} Reduction is relative to $T\times D$.\par}
\end{table}

\section{Runtime on Additional Datasets}
\label{app:runtime_scaling}
This appendix reports runtime results on the remaining benchmark datasets.

\begin{table}[H]
\centering
\caption{Runtime at $M{=}50$.}
\label{tab:runtime_additional}
\setlength{\tabcolsep}{4pt}
\renewcommand{\arraystretch}{0.95}
\scriptsize
\begin{tabularx}{\linewidth}{l *{3}{>{\centering\arraybackslash}Y}}
\hline
Method & HAR & PTB-XL & S\&P500 \\
\hline
KernelSHAP   & $0.022 \pm 0.003$ & $\mathbf{0.083 \pm 0.007}$ & $\mathbf{0.027 \pm 0.002}$ \\
TimeSHAP     & $0.281 \pm 0.020$ & $29.721 \pm 0.342$ & $2.813 \pm 0.034$ \\
SequenceSHAP & $0.584 \pm 0.071$ & $3.002 \pm 0.165$ & $1.660 \pm 0.025$ \\
WindowSHAP   & $0.182 \pm 0.015$ & $0.247 \pm 0.020$ & $0.309 \pm 0.021$ \\
TSHAP        & $0.201 \pm 0.017$ & $0.281 \pm 0.022$ & $0.352 \pm 0.024$ \\
GS-SHAP      & $\mathbf{0.176 \pm 0.023}$ & $2.381 \pm 0.143$ & $1.194 \pm 0.078$ \\
\hline
\end{tabularx}
\vspace{2pt}
\footnotesize
\noindent{\raggedright \textbf{Note.} Mean $\pm$ std. seconds/sample over 100 samples. Lower is faster.\par}
\end{table}

\FloatBarrier
\begin{acks}
This work was supported by the National Research Foundation of Korea (NRF) grant funded by the Korea government (MSIT) (No. RS 2025-00554384), and the Technology Development Program (No. RS 2024-00513926) funded by the Ministry of SMEs and Startups (MSS, Korea).
\end{acks}

\section*{GenAI Usage Disclosure}
This paper includes text that was revised and refined using generative AI tools (ChatGPT) to improve clarity and correct potential grammatical errors. All scientific ideas, experiments, analyses, and contributions are the authors' own and were reviewed and verified by the authors.

\bibliographystyle{ACM-Reference-Format}
\balance
\bibliography{references}

\end{document}